%% file: main.tex
\pdfoutput=1
\documentclass[11pt]{article}

\usepackage[final]{acl}

\usepackage{times}
\usepackage{latexsym}
\usepackage{makecell}
\usepackage{array}
\usepackage{longtable}
\usepackage{graphicx}
\usepackage{booktabs}
\usepackage{multirow}
\usepackage{multicol}
\usepackage{kotex}
\usepackage{colortbl} % 테이블에 색을 넣기 위한 패키지
\usepackage[linesnumbered,ruled]{algorithm2e}
\usepackage{amsmath, amsfonts,amssymb}
\usepackage{MnSymbol}
\usepackage{verbatim}
\newcommand{\red}[1]{\textcolor{red}{#1}} %%%% remove!!!
\usepackage{pifont}
\newcommand{\cmark}{\ding{51}} % Checkmark
\newcommand{\xmark}{\ding{55}} % Crossmark
\usepackage{subcaption}
\usepackage{soul}
\usepackage{hyperref}

\usepackage[T1]{fontenc}
\usepackage[utf8]{inputenc}

\usepackage{microtype}

\usepackage{inconsolata}

\title{
DCG-SQL: Enhancing In-Context Learning for Text-to-SQL\\with Deep Contextual Schema Link Graph
}
\author{Jihyung Lee$^{1}$\thanks{Equal contribution}, Jin-Seop Lee$^{1}$\footnotemark[1], Jaehoon Lee$^{1}$, YunSeok Choi$^{2}$\thanks{Corresponding author}, Jee-Hyong Lee$^{3}$\footnotemark[2]\\
$^{1}$Department of Artificial Intelligence, Sungkyunkwan University \\
$^{2}$Department of Immersive Media Engineering, Sungkyunkwan University \\
$^{3}$Department of Computer Science and Engineering, Sungkyunkwan University \\
  \texttt{\{jjklle, wlstjq0602, hoon1223, ys.choi, john\}@skku.edu}}

\begin{document}
\maketitle
\begin{abstract}
Text-to-SQL, which translates a natural language question into an SQL query, has advanced with in-context learning of Large Language Models (LLMs). However, existing methods show little improvement in performance compared to randomly chosen demonstrations, and significant performance drops when smaller LLMs (e.g., Llama 3.1-8B) are used. This indicates that these methods heavily rely on the intrinsic capabilities of hyper-scaled LLMs, rather than effectively retrieving useful demonstrations. In this paper, we propose a novel approach for effectively retrieving demonstrations and generating SQL queries. We construct a Deep Contextual Schema Link Graph, which contains key information and semantic relationship between a question and its database schema items. This graph-based structure enables effective representation of Text-to-SQL samples and retrieval of useful demonstrations for in-context learning. Experimental results on the Spider benchmark demonstrate the effectiveness of our approach, showing consistent improvements in SQL generation performance and efficiency across both hyper-scaled LLMs and small LLMs. The code is available at \href{https://github.com/jjklle/DCG-SQL}{https://github.com/jjklle/DCG-SQL}.
\end{abstract}

\input{texs/1.intro}

\input{texs/2.related}

\input{texs/3.method}

\input{texs/4.experiment}

\input{texs/5.conclusion}
\bibliography{custom}

\input{texs/6.appendix}

\end{document}

%% file: texs/1.intro.tex
\section{Introduction}
Text-to-SQL aims to generate SQL queries given a pair of a natural language question and a database schema. 
Traditional studies focused on how to train models that took a question and a schema as input to generate the target SQL \cite{RAT-sql,RASAT,graphix}. However, they required training an encoder-decoder model with billions of parameters, leading to high training costs and limited generalization. Recently, with the advance of Large Language Models (LLMs), in-context learning with LLMs has been applied to Text-to-SQL tasks due to its efficiency and strong generalization ability across various tasks~\cite{EvaluateText-to-SQL,howtoprompt,DIN}.

\begin{table}
\centering
\small
% \resizebox{0.8\linewidth}{!}
\begin{tabular}{lc|cc}
\toprule
Methods & Retrieval & GPT-4 & Llama 3.1-8B  \\ \toprule
% DIN-SQL & -- & 82.8  & \hl{00.0} \\ \midrule
\multirow{2}{*}{ACT-SQL} & \cmark & 83.9  & 75.0 \\ 
& \xmark & 83.2 & 75.6 \\ \midrule
\multirow{2}{*}{DAIL-SQL} & \cmark & 82.1  & 69.9   \\
& \xmark & 80.8  & 69.3    \\
% \bottomrule \rule{0pt}{1.0EM}
% DCG-SQL & \textbf{87.5} & \textbf{82.1} \\
\bottomrule
\end{tabular}
\vspace{-0.2cm}
\caption{Execution Accuracy (\%) for Spider Dataset. 
All experiments are conducted with 4-shot setting.
When \xmark\ is applied, the demonstrations are randomly choosen.
}
\label{intro_table}
% \vspace{-0.6cm}
\end{table}

To maximize the capability of in-context learning of LLMs, it is necessary to provide relevant and useful demonstrations to generate correct SQL query~\cite{in-context}. 
Most existing approaches represent each sample using a text embedding derived solely from the question, and select demonstrations based on the similarity of the question embeddings~\cite{MCS,GPT-3.5, ACT,DAIL}. However, as shown in Table \ref{intro_table}, their performance shows no difference from that of randomly chosen demonstrations. This result implies that existing methods failed to maximize the capabilities of in-context learning due to their limitation in selecting appropriate demonstrations. This becomes more apparent when they are applied to smaller LLMs which have weaker in-context learning capabilities compared to hyper-scale LLMs~\cite{emergent}. Since smaller LLMs rely more heavily on demonstrations, performance significantly drops, sometimes even falling below the performance with randomly selected demonstrations.

This is because they overlook DB schema when representing Text-to-SQL samples. Since DB schema determines the structure and semantics of SQL queries, it must be incorporated into the representation. In Text-to-SQL tasks, understanding a question inherently relies on the database schema. Therefore, when retrieving relevant demonstrations, both the question and the schema should be jointly considered and represented in a unified manner. encoded through a unified representation.
% When retrieving demonstrations, we need to consider both the question and the schema. They need to be represented in a unified manner, rather than simply concatenating their embeddings in parallel.

In this paper, we propose a novel Text-to-SQL method for in-context learning with \textbf{D}eep \textbf{C}ontextual Schema Link \textbf{G}raph-based Retrieval, as we called DCG-SQL.
We develop a novel graph-based joint representation of both the question and the schema to retrieve demonstrations.
% By utilizing this representation, we select demonstrations that exhibit a high contextual similarity between the given question and the schema, to maximize the reasoning capability of LLMs. 
%Our approach significantly enhances performance not only for hyper-scale LLMs but also for smaller LLMs, demonstrating substantial improvements across various model scales.
However, developing a joint representation that effectively captures the contextual relationship between the question and the schema presents two challenges.

First, datasets are not usually available to learn how schema items are relevant to the question, a process commonly known as schema linking~\cite{old_schema_linking1, RAT-sql}. As a result, some previous works~\cite{RASAT, graphix, REFSQL} have attempted to establish links through simple word matching, but this approach is often too simplistic and fails to accurately capture the relationship between questions and schema items. A novel method is needed to effectively model the contextual relationships necessary for Text-to-SQL samples without annotations.

The second challenge is not all schema items in a database are necessary for effectively representing the given Text-to-SQL sample. If all schema elements are considered, irrelevant schema items could hinder the effective representation.
If irrelevant schema items are included in representations, question tokens could be connected to unrelated schema items.
They also make it difficult to distinguish between samples with different questions that share the same DB schema.
Therefore, when representing Text-to-SQL samples for demonstration retrieval, it is also essential to exclude irrelevant schema items.

To identify only relevant schema items to the question in a sample, we train a transformer based encoder with classification heads. Since each training sample contains a question, a schema and a target SQL, we can train the classifier with the question and the schema as input, and the SQL to indicate relevant schema items. When training the encoder, we add special tokens, such as [TAB] and [COL], to generate appropriate representations even when their names are composed of several words. These special tokens help to effectively capture their semantics. Next, we generate a graph with the selected schema items and the words in the question. We connect items and words using attention scores between question tokens and schema item tokens. Since the attention mechanism enables the model to assign relevance between tokens~\cite{attention}, we use the attention scores to predict the schema linking. This allows the model to capture the contextual relationships without any explicit linking dataset. With these, we can effectively achieve a joint representation of each text-to-SQL sample as a graph. By comparing the representations of the graphs, we can retrieve demonstrations. With the demonstrations, our method fully leverages in-context learning capability and generate the target SQL correctly.

To evaluate our proposed method, we conduct experiments on various benchmark datasets, including Spider, Spider-DK, Spider-Realistic, and Spider-Syn~\cite{spider,dk,realistic,syn}.
Moreover, we experiment with various hyper-scaled and small LLMs, including GPT-4, GPT-3.5-Turbo, DeepSeek-Coder-33B, Llama 3.1-8B, DeepSeek-Coder-6.7B, and Llama 3.2-3B~\cite{GPT-4,Deepseek,llama3}. Our method consistently achieves performance improvements not only with hyper-scaled LLMs but also with small LLMs.

%% file: texs/2.related.tex
\section{Related Work} \label{2}

% \subsection{LLM centric In-context learning for Text-to-SQL}
\begin{comment}
LLM을 이요한 in-context learning은 
Evaluating the text-to-sql, How to Prompt, exploring the chain of thought <- 고정 샷
Din, pet, MCS <<=== LLM 의존 작은거 쓰면 떨어짐
act, gpt 3.5 question masking < === 의존은 낮은데, SQL 특징 반영못함
Enhancing fewshot, dail <--- SQL 구조 다 못봄

MCS -- question 유사도, question mask 유사도 retrieve
PET -- question mask 유사도 retrive

GPT 같은 큰거 ---> demonstraion에 영향을 덜받음. 기본 성능 높게나옴(제로샷 혹은 random fewshot해야하나??)
\end{comment}

\begin{figure*}[t]
  \includegraphics[width=\linewidth]{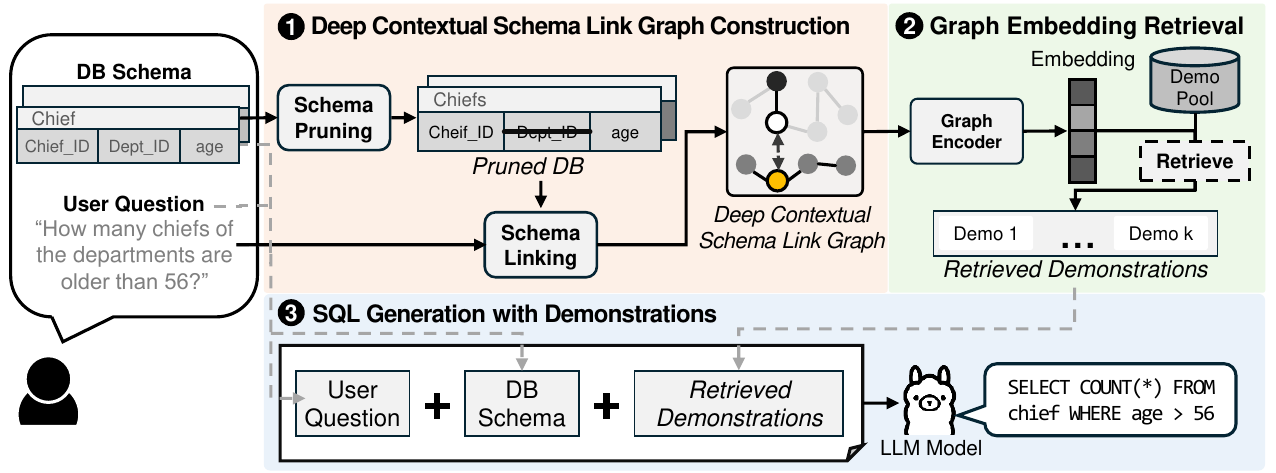} 
  \caption {\textbf{Overview of our proposed method for Text-to-SQL.} The method consists of three main steps: (1) Deep Contextual Schema Link Graph Construction, which prunes irrelevant schema items and
  builds a schema link graph that captures the contextual relationship between the question tokens and the schema items; (2) Graph-based Demonstration Retrieval, where graph embeddings are used to retrieve demonstrations relevant to the given query; and (3) SQL Generation using retrieved demonstrations to generate the final SQL query.}
    \label{mainfig}
% \vspace{-0.2cm}
\end{figure*}

% \paragraph{GPT-based Text-to-SQL Approaches}
% Conventional Text-to-SQL research train the encoder-decoder models to generate SQL queries from input questions~\cite{sqlnet, seq2sql,syntaxsqlnet}. 
% \subsection{In-context Learning for Text-to-SQL}
% \red{Conventional Text-to-SQL research utilizes schema graph to effectively extract representations for input questions and DB schemas~\cite{RASAT, RAT-sql, REFSQL, graphix, SADGA}. When constructing a graph, they involve all DB schemas and make links only when DB schema items exactly match question tokens.
% However, as a result, they include irrelevant DB schemas for the target SQL query and fail to capture semantic relationships in different forms, leading to an ineffective graph. Moreover, since they are based on encoder-decoder models to generate SQL queries, they require substantial computational cost to train the decoder model and also demonstrate lower generalization capability compared to LLMs.}
\subsection{In-context Learning for Text-to-SQL}

With the advancement of large language models, there has been a growing focus on leveraging them to solve Text-to-SQL tasks \cite{DIN, howtoprompt,GPT-3.5,ACT,MCS, DAIL, Enhancing,PET, ast}. Some approaches decompose the Text-to-SQL task into multiple steps and employ GPT-4 at each step \cite{DIN, PET, MCS}. However, this repeated use of GPT-4, a hyper-scaled LLM, results in significantly high inference costs. 
Also, since they heavily depend on the hyper-scaled LLM's capabilities, their performance significantly drops with smaller LLMs, as limited capability of small LLM at each step leads to error accumulation throughout the process.

On the other hand, others \cite{ACT, GPT-3.5, Enhancing, DAIL, ast, PET, MCS} tried to leverage in-context learning capability of LLMs by retrieving relevant samples and providing them as demonstrations.
\citet{ACT, GPT-3.5, PET, MCS} encoded the question into a text embedding using a pre-trained text encoder and retrieved demonstrations based on the embedding similarity. However, they did not consider the DB schema, which is also crucial in Text-to-SQL.
In contrast, \citet{DAIL, Enhancing, ast} tried to reflect DB schema information indirectly by approximating target SQL of user input and retrieving similar examples through similarity comparison of the approximated SQL. However, they still have the limitation of requiring costly training and relying on the approximator which has limited capacity for generating SQL. As a result, inaccurate DB schema information can be represented when the approximator generates wrong SQL.
Since they did not effectively consider the DB schema during retrieval, they failed to provide useful demonstrations for LLMs' in-context learning.

% In addition, an error in any step can propagate through subsequent steps, consequently resulting in incorrect outcomes. 

% Also, when smaller LLMs with limited in-context learning capability are used, they tend to make more errors at each step, further degrading performance, as shown in Table \ref{intro_table}.
% Also, since smaller LLMs have limited in-context learning capability, they tend to make more errors at each step which further degrades performance, as shown in Table \ref{intro_table}. 
%%% 이 얘기는 안해주어도 되지않을까?

% \cite{ACT, GPT-3.5, PET, MCS}는 pre-trained 된 text-encoder를 통해 각 question의 embedding으로 각 sample을 표현하였고, 이 embedding의 유사도 비교를 통해 유사한 demonstration을 retrieval 하였다. \cite{DAIL, Enhancing, ast}는 tried to reflect DB schema information indirectly by approximating target SQL of test case and retrieving through similarity comparison of the approximated SQL. However, they have a limitation of embedding을 표현하는데 DB 정보를 효과적으로 incorporate 하지 못하거나, relying on the approximator which has limited capacity for generating SQL. As a result, inaccurate DB schema information can be represented when the approximator generates wrong SQL. 

% This leads to retrieving irrelevant demonstrations. In addition, training the approximator with 3B parameters, such as Graphix-T5 \cite{graphix}, also requires substantial computational cost. 
% As a result, as shown in Table \ref{intro_table}, the performance of smaller model drops significantly.

\subsection{Graph representation for text-to-SQL samples}
Before the advent of LLMs, Text-to-SQL research learned schema linking information, which explicitly associates question tokens with relevant database schema elements, to generate target SQL using encoder-decoder models. This helps the model interpret the question and recognize how schema components are organized within the database, leading to improve SQL generation accuracy \cite{schemalinking}.

To provide schema linking information to the model, \citet{RAT-sql, RASAT, REFSQL, graphix} represented the input in a graph format, reflecting the relationships between question tokens and database schema elements. Due to the lack of datasets that explicitly annotate schema linking, these methods were based on simple textual matching, comparing only the spelling between question tokens and schema items. However, such simple linking often fails to capture deeper contextual relationship when question token and schema items are expressed in different forms.

% For example, consider the question, “How many heads of the department are older than 56?” and a given database schema with a table named 'heads' consisting of the columns 'name' and 'age'. They can make a link between the question token ‘heads’ and table ‘heads’ based on spelling match. However, they cannot capture the relationship between ‘age’ column and the question tokens ‘older’ and ‘56’ as they are in different form.

Furthermore, their graph representations contain all schema items, even those unrelated to the question, which can lead to incorrect schema linking. Moreover, retrieval becomes challenging when all schema items are included, as each graph is mostly composed of DB schema elements, making samples that share the same DB schema indistinguishable. Therefore, to retrieve useful demonstrations, a more effective graph representation is needed—one that captures deeper contextual relationships while including only the relevant schema elements.

%% file: texs/3.method.tex
\begin{figure*}[t]
  \includegraphics[width=\linewidth]{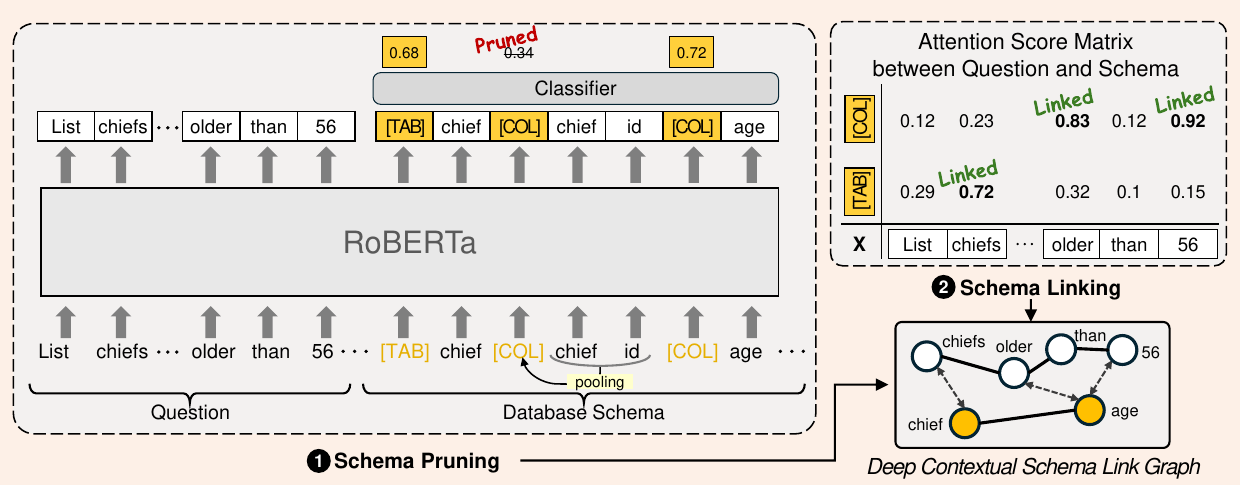}
  \caption {Deep Contextual Schema Link Graph Construction.}
    \label{figure3}
% \vspace{+0.cm}
\end{figure*}

\section{Proposed Method}
% In this section, we introduce a novel graph representation to effectively retrieving useful demonstrations for in-context learning in Text-to-SQL task.
In this section, we introduce a Text-to-SQL method with \textbf{D}eep \textbf{C}ontextual Schema Link \textbf{G}raph-based Retrieval, DCG-SQL.
Figure~\ref{mainfig} illustrates an overview of our method, consisting of three main processes: Deep Contextual Schema Link Graph Construction, Graph-based Demonstration Retrieval, and SQL Generation.

First, we construct a deep contextual schema link graph, which captures the contextual relationships between question and relevant schema items.
Then, to retrieve demonstrations with our graph representation, we train a graph encoder with self-supervised learning. 
This ensures that samples closer to the anchor are more useful in generating the target SQL, enabling the effective retrieval of useful demonstrations.
Finally, with the selected demonstrations, our method can fully leverage in-context learning ability, and generate the target SQL more correctly.

%In this section, we introduce a novel approach for effectively retrieving demonstrations and generating SQL queries using small LLM. %generating SQL

 % 우리는 Text-to-SQL에서 효과적인 In-context learning을 위해, DB schema와 question를 동시에 고려하여 few shot example을 retrieve하는 QS-retrieval 방법을 제안한다. 이 방법은 question token들과 db schema의 정보를 동시에 담은 QS representation을 구성하는 단계와, 이를 encoding 하는 dense retriever를 통해 few shot example을 선택하는 단계로 이루어져 있다. 이 같은 과정은 section 3.1과 section 3.2에 나타난다. 또한, In-context learning의 능력이 상대적으로 제한적인 sLLM에서 효과적인 target SQL 생성을 위한 Schema Item selection 및 automated CoT 방법을 제안한다. 이러한 과정은 section 3.1과 section 3.3에 나타난다.

% 이를 이용하여, triplet을 하나의 embedding으로 encoding 하는 dense retriever를 학습한다. 학습된 retriever를 이용하여, user question과 user db schema와 유사한 demonstration들을 선택하고, 선택된 example들을 fewshot으로 하여 In-context learning을 진행한다.

% Schema Pruning 방법은 DB schema에서 target SQL query 생성에 필요하지 않은 table과 column들을 제거하는 방법이다. 두번째는, DB schema와 question 그리고 그 둘 사이의 linking을 고려하여 도움이 되는 demonstration을 선택하는 방법이다. 세번째는 automated-cot를 활용한 generation 방법, 그리고 마지막으로 Few shot sample revision 방법이다.

% domain-adapted 기반의 RaR 방법을 제안한다. 이 방법은 두 단계로 구성되며, 첫번째로 데이터베이스의 구조와 질문의 의미적 유사도를 동시에 고려하는 retrieval step, 두번째로는 retrieval된 sample을 test case에 도메인에 맞게 변환하는 recosntruction이다. 두 과정을 통해 test case와 동일한 도메인의 few shot sample을 생성하여 SQL 생성 능력을 향상시킬 수 있다. RaR의 전체적인 framework는 \ref{mainfig}에 나타내었다.

\subsection{Deep Contextual Schema Link Graph Construction} % Construction
%Inputting all schema items as plain text during in-context learning causes the model to miss essential schema components. 
%To better model the schema structure that contains important components,
We propose a deep contextual schema link graph construction method that excludes irrelevant schema items and extracts only the relevant ones. Our method also links the question tokens with the schema items based on their contextual relation.

% DB schema와 question을 모두 반영하는 embedding을 위해서, text-to-SQL sample(question, DB schema)을 하나의 Schema Link Graph Construction표현한다. \cite{Rasat, rat-sql, REFSQL} 이때, 모든 schema item을 graph의 반영하는 경우, text-to-SQL sample을 효과적으로 나타낼 수 없다. 또한, question token과 schema item을 link를 통해, question이 요구하는 정보를 효과적으로 파악하게 한다.
% % rasat, rat-sql, ref-sql과 같이 하나의 text-to-SQL sample을 (question token들과 DB schema item들을 노드로 하고, 그들 사이의 relation을 edge로 하는) 그래프로 나타낼 수 있다.\site{rasat} 하지만 이 방법은 SQL 생성에 불필요한 schema item들을 다수 포함하고 있으며, question token과 DB schema간의 관련성을 제대로 파악하지 못해, sample을 효과적으로 표현할 수 없다.
% 따라서 우리는 schema pruning을 통해 주요한 schema item만을 extract하고, schema linking 방법을 통해 question과 schema item의 관계를 반영한 그래프를 구성하는 방법을 제안한다.

\paragraph{Schema pruning} DB schema consists of multiple tables and columns, but only a small subset is relevant for SQL generation.
If irrelevant DB schema items are included in the graph, they could hinder extracting representation and retrieving useful demonstrations. To focus on the relevant components, we introduce a schema pruning method that identifies the relevant schema items necessary for generating the SQL.
%To focus on these essential components, we introduce a schema pruning method that identifies the relevant schema items necessary for generating the SQL.

For schema pruning, we utilize a cross-encoder model to determine the relevance of tables and columns \cite{RESDSQL}. 
The input to the cross-encoder is a flattened text sequence that combines the question with schema items (either a table or a column). 
Since schema items, such as the column name `head id', may be tokenized into multiple tokens, they need to be represented as a single token. 
We prepend special tokens [TAB] and [COL] for representing tables and columns, respectively.
The special token embeddings for schema items are initialized by applying mean pooling over the token embeddings of each word in the schema item. Unlike existing pruning approach \cite{RESDSQL}, these special tokens enable us to leverage the attention scores between question and schema items.

Let $Q$ denote the tokens of the question, and $T$ and $C$ denote the names of the table and column. 
The input sequence $X$ is represented as: 
\begin{align}
X &= \{ Q \, \textnormal{[TAB]} \, T_1 \, \textnormal{[COL]} \, C_{11}, \ldots, \nonumber \\
& \textnormal{[COL]} \, C_{1n} \, \textnormal{[TAB]} \, T_2 \, \textnormal{[COL]} \, C_{21}, \ldots \} \nonumber
\end{align}
where $T_i$ denotes the $i$-th table in the DB schema, and $C_{ij}$ refers to the $j$-th column of the table $T_i$.
The cross-encoder model is trained as binary classification for each special token to predict the relevant tables and columns for a given question, as shown in Figure \ref{figure3}.

% DB schema는 보통 다수의 table과 column들로 구성되어 있지만, question 및 target SQL과 관련된 table 및 column은 소수에 불과하다. LLM의 demonstraction을 위하여 question에 대한 불필요한 schema item들을 포함하면 불필요한 정보들이 많아진다. 이를 해결하기 위해, target SQL 생성에 필요한 SQL-relevant schema item만을 identify하는 schema pruning 과정을 진행한다.

% For schema pruning, 우리는 cross-encoder 모델을 기반으로 table과 column의 사용 여부를 classify한다. 이 cross-encoder 모델의 입력은 question과 table, column을 모두 text로 표현하고 flatten한 sequence로 구성한다. 이때, One schema item could be tokenized into multiple tokens (i.e. column name 'card type code').
% 이러한 item의 classification을 위해서는 이를 one token representation으로 표현해야 한다. 우리는 table과 column을 나타내는 [TAB], [COL] token을 정의하여 각 schema item 앞에 concat하고, schema item의 word embedding들을 mean pooling하여 word embedding을 intialize한다. 

% 이러한 방법은 schema item과 question token간의 schema linking에 이용할 수 있다. 이와 같은 schema item selection 모델의 input X은 다음과 같다.
% q denotes question token, t and c denotes name of the table and column. | is delimiter.
%    X= q1,q2,q3,...,qn | [TAB] t1: [COL] c11,c21, ..., cn1|

   % Unlike \cite{RESDSQL}, attached a pooling module consist of Bi-LSTM layers after cross encoder,

% Identifying schema items contextually relevant to the question is crucial for understanding the information required by the question and generating accurate SQL outputs \cite{schemalinking, graphix,RAT-sql, REFSQL}. 
\paragraph{Schema linking}
Identifying schema items contextually relevant to the question is crucial for understanding the question and generating SQL outputs \cite{schemalinking, graphix, RAT-sql}. 
However, due to the absence of ground truth dataset for such relation, existing approaches rely on simple text correspondence \cite{graphix, REFSQL,RASAT,RAT-sql}. It compares only the spelling between question tokens and schema items, often failing to capture links when schema items are phrased differently (e.g., linking the question tokens "\textit{older}" and "\textit{56}" to the column "\textit{age}" in Figure \ref{figure3}).
%Therefore, we perform schema linking to construct connections between the question and the pruned schema items based on their correspondence (either exact match or partial match) \cite{RAT-sql,RASAT}.
%Furthermore, to handle cases where schema elements and question tokens differ in form but share similar meaning, we connect links based on their semantic similarity.
% Furthermore, to handle cases where  schema items and question tokens have similar meanings but differ in form or wording, we establish links based on their semantic similarity.x` 이거 넣어야 뒷 내용 연결 될것 같음. 아니면 뒷 문장 수정해야... 근데 다시보니 필요없을거 같기도..

%To capture semantic similarities between tokens, we leverage the attention scores from the trained cross-encoder model.

To overcome this issue and capture contextual similarities, we leverage the attention scores from the schema pruning model. Since the pruning model is trained to identify relevant schema items for a given question, its attention scores implicitly reflect the relevance between them. 
%%%% 이거 어디에다가 넣어야 하는 거예요?
% Unlike existing pruning approach \cite{RESDSQL}, we use the special tokens that encapsulate the information of each schema item, enabling us to leverage the attention scores of the pruning model. 
%%%% 이거 어디에다가 넣어야 하는 거예요?
If the schema pruning model classifies a schema item as relevant to SQL generation, we obtain the attention scores between its concatenated special token and the question tokens. 

When the attention score exceeds the threshold $\tau$, a link is connected between the schema item and the question token, which we refer to as an `attention-match'. 
%If a schema item is classified as relevant to SQL generation, we obtain the attention scores between its concatenated special token and the question tokens. When the attention score exceeds a threshold $\tau$, a link is connected between the schema item and the question token, which we refer to as an 'attention-match'. 
This approach enables schema pruning and linking simultaneously without ground truth dataset or additional training.
%Using the trained pruning model, schema linking is performed easily through attention scores without any additional training.
% For token-to-token edges, to reflect the grammatical structure of the question, we link tokens that are adjacent or have high syntactic correlation. For schema-to-schema edges, to incorporate the structure information of the database, we link all schemas within each table.
Also, we link tokens that have high syntactic correlation for edges between question and all schema items within each table for edges between schema items, as following existing approaches \cite{RASAT, REFSQL, graphix, RAT-sql}.
% \red{For edges between question tokens and edges between schema items, we link tokens that are adjacent or have high syntactic correlation and all schemas within each table, respectively ,as following follow existing approaches \cite{RASAT, REFSQL, graphix}}
% \red{그리고, For links between question tokens and links between schema items, we follow existing approaches \cite{RASAT, REFSQL, graphix}. For token-to-token edges, to reflect the grammatical structure of the question, we link tokens that are adjacent or have high syntactic correlation. For edges between schema items, we make link based on the structural information in a DB schema such as PRIMARY-KEY, FOREIGN-KEY, BELONGS-TO or SAME-TABLE etc.}
% We can identify semantic similarities and create links without additional model training.

Finally, we construct a deep contextual schema linking graph $G$, where question and schema are jointly represented. The graph consists of nodes, which correspond to question tokens and schema items, and edges that capture the contextual relationships between them.

\subsection{Graph-based Demonstration Retrieval}
\label{sec:3.2}
%%Existing retrieval methods \cite{PET,MCS,ACT} typically search for similar demonstrations based on only natural language questions. However, in Text-to-SQL, both the question and DB schema are important in generating SQL queries. Therefore, both must be considered during demonstration retrieval. 

% \begin{figure}[t]
%   \includegraphics[width=\linewidth, height=1.0\linewidth]{figures/fig3.pdf}
%   % \vspace{-0.7cm}
%   \caption {Graph-based Demonstration Retrieval \red{재훈 그림 수정 필요}}
%     \label{figure3_2}
% \end{figure}

%%We propose a graph-based retrieval method that leverages schema link graph with both the question and the DB schema to retrieve relevant demonstrations.
To retrieve useful demonstrations based on the deep contextual schema link graph, we propose a graph-based retrieval method that jointly consider the question and the DB schema.
%%We adopt a Relation-Aware Transformer (RAT) encoder as a retrieval model to embed each sample as a schema link graph \cite{relationtransformer, RASAT}.
To achieve this, we adopt a  graph encoder to effectively represent each sample's graph into an embedding vector~\cite{RAT-sql,RASAT}.

%%%%% RAT 에 대한 설명
\begin{comment}
This encoder incorporates relation-aware self-attention, where each token's relational information is embedded into the self-attention mechanism using relation embeddings. Specifically, the input consists of the question and the corresponding DB schema represented as text, along with a relation matrix which encodes the relations between question tokens and schema items. 
The encoder uses \red{mean pooling} to convert the schema link graph into an embedding vector, effectively capturing both the question and the DB schema.
\end{comment}
%%%%% RAT 에 대한 설명

% To effectively train schema link graph embeddings for retrieval, 
To effectively train the graph encoder, we utilize contrastive learning that clusters similar graphs closely in the vector space while distancing dissimilar ones farther apart. 
We define positive demonstration samples as those that are useful for generating the SQL query of a given anchor. To achieve this, for the given anchor sample $x = (G_x, SQL_x)$,
% \red{we first rank other samples $S = \{s_i=(G_i, SQL_i) \mid i = 1, 2, \dots\}$ based on SQL similarity, using the tree-edit-score between $SQL_x$ and $SQL_i$.
we extract likelihoods of the $SQL_x$ when other samples  $S = \{s_i=(G_i, SQL_i) \mid i = 1, 2, \dots\}$ are provided as demonstrations using Llama 3.1-8B model~\cite{llama3}. The likelihood is as follows:
% Then we extract likelihoods of the $SQL_x$ when other samples $s_i$ are provided as demonstrations using Llama 3.1-8B. The likelihood is as follows:}
\begin{equation}
score(x,s_i) = P_{\text{sLLM}}(SQL_x \mid G_x ; s_i)
\end{equation}
If the likelihood of $SQL_x$ is high, it indicates that $s_i$ is useful for generating the target SQL query.

To identify positive and negative samples, we rank $s_i$’s based on their likelihood of generating the anchor's target SQL, treating high-ranked ones as positives and low-ranked ones as negatives. 
% \red{This process enables hard negative mining by filtering out samples that looks similar but are not actually useful for generating the target SQL.}
Since the retrieval model is trained to increase similarity with high-ranked demonstrations and decrease similarity with low-ranked ones, we can retrieve useful demonstrations from the retrieval pool through similarity of the graph embeddings~\cite{dpr, udr}.

However, computing the likelihood of $SQL_x$ over the entire training dataset as demonstrations is computationally expensive. To address this, we construct a reduced candidate set for each anchor by selecting 100 structurally similar SQLs, using a tree edit distance computed over abstract syntax trees~\cite{syntaxsqlnet}. Unlike \citet{ast}, which selects demonstrations based on AST similarity, we further re-rank the candidates by their usefulness for generating the anchor SQL. This enables us to identify samples that appear structurally similar but are unhelpful as demonstrations—performing hard negative mining.

During inference, we compare the embedding of user input graph $G_{user}$ generated by the encoder, with the embeddings of the demonstration pool by computing the similarity. We select top-k samples with the highest similarity as demonstrations.

\subsection{SQL generation with In-context Learning}
With the retrieved demonstrations, LLM can generate target SQL query correctly. In addition, we utilized automated-CoT prompting method to guide the SQL generation more effectively~\cite{automated-cot, ACT}. The retrieved demonstrations are categorized into types such as simple, join, and nested queries. 
Each type uses a predefined CoT template, dynamically adapted per demonstration to guide consistent and structured SQL generation. Details are provided in Appendix~\ref{appendix:automated-cot}.

%% file: texs/4.experiment.tex
\begin{table*}[t]
\centering
% \Large
\small
% \scriptsize
%\renewcommand{\arraystretch}{1.1}
%\resizebox{\textwidth}{!}{

\renewcommand{\arraystretch}{1.1}

\begin{tabular}{lcccccccccc}
\toprule
\multicolumn{1}{c}{\multirow{2}{*}{\begin{tabular}[c]{@{}c@{}} 
 \textbf{Method}\vspace{-5pt}\end{tabular}}} &
\multicolumn{2}{c}{\textbf{Easy}} & \multicolumn{2}{c}{\textbf{Medium}} & \multicolumn{2}{c}{\textbf{Hard}} & \multicolumn{2}{c}{\textbf{Extra}} & \multicolumn{2}{c}{\textbf{All}} \\ \cmidrule{2-11}
\multicolumn{1}{c}{} & 
\multicolumn{1}{c}{EX} &
\multicolumn{1}{c}{EM} &
\multicolumn{1}{c}{EX} &
\multicolumn{1}{c}{EM} &
\multicolumn{1}{c}{EX} &
\multicolumn{1}{c}{EM} &
\multicolumn{1}{c}{EX} &
\multicolumn{1}{c}{EM} &
\multicolumn{1}{c}{EX} &
\multicolumn{1}{c}{EM} \\
\bottomrule
\rowcolor{gray!25}\multicolumn{11}{c}{Finetuned Model}  \\

Graphix-T5   & 92.3 & 91.9 & 86.3 & 82.3 & 73.6 & 65.5 & 57.2 & 53.0 & 80.9 & 77.1 \\
RESD-SQL     & - & - & - & - & - & - & - & - & 84.1 & 80.1 \\
\bottomrule
\rowcolor{gray!25}\multicolumn{11}{c}{GPT-4}  \\
DIN-SQL   & 92.3 & 82.7 & 87.4 & 65.5 & 76.4 & 42.0 & 62.7 & 30.7 & 82.8 & 60.1 \\
ACT-SQL   & 92.3 & 85.9 & 89.3 & 59.1 & 80.2 & 53.1 & 61.5 & 18.9 & 83.9 & 57.9 \\
DAIL-SQL   & 91.1 & 90.3 & 88.6 & 76.0 & 75.9 & 60.3 & 62.0 & 50.0 & 82.8 & 72.6 \\
ASTRES &-&- & -& -& - & - & - &-& 86.6 &77.3 \\
DCG-SQL (Ours)   & \textbf{95.6} & \textbf{92.3} & \textbf{90.6} & \textbf{83.0} & \textbf{82.8} & \textbf{70.7} & \textbf{72.3} & \textbf{61.4} & \textbf{87.5} & \textbf{79.7} \\
\bottomrule
\rowcolor{gray!25}\multicolumn{11}{c}{GPT-3.5-Turbo}  \\
DIN-SQL   & 85.1 & 56.5 & 81.8 & 51.6 & 62.1 & 28.2 & 51.2 & 7.8 & 74.4 & 41.8 \\
ACT-SQL   & \textbf{92.7} & 87.9 & 86.1 & 52.5 & 70.1 & 41.4 & 56.6 & 13.3 & 80.3 & 52.8 \\
DAIL-SQL   & 90.3 & \textbf{89.9} & 82.7 & 57.0 & 73.0 & 52.3 & 56.0 & 18.7 & 78.0 & 64.9 \\
ASTRES &-&-&-&-&-&-&-&-&83.0&68.8\\
DCG-SQL (Ours) &91.9&88.3&\textbf{89.2}&\textbf{78.0}&\textbf{74.7}&\textbf{56.3}&\textbf{63.3}&\textbf{39.8}&\textbf{83.3}&\textbf{70.7} \\
\bottomrule
\rowcolor{gray!25}\multicolumn{11}{c}{Deepseek-Coder-33B-Instruct}  \\
DIN-SQL   & 91.9 & 81.9 & 82.1 & 62.3 & 58.0 & 31.0 & 50.0 & 17.5 & 75.0 & 54.5 \\
ACT-SQL   & 84.3 & 78.6 & 85.0 & 57.4 & 64.4 & 44.8 & 50.6 & 25.3 & 75.8 & 55.2 \\
DAIL-SQL   & 90.3 & 87.9 & 86.5 & 73.8 & 73.0 & 62.6 & 57.2 & 45.2 & 80.5 & 70.7 \\
ASTRES &-&-&-&-&-&-&-&-&83.4&64.7 \\
DCG-SQL (Ours)   & \textbf{94.0} & \textbf{90.7} &\textbf{92.6}& \textbf{80.0}&\textbf{75.3}&\textbf{63.8}&\textbf{63.3} & \textbf{53.0} & \textbf{85.3} & \textbf{75.5} \\
\bottomrule
\rowcolor{gray!25}\multicolumn{11}{c}{Llama 3.1-8B-Instruct}  \\
DIN-SQL   & 84.7 & 79.4 & 72.2 & 54.9 & 60.9 & 36.8 & 35.5 & 19.9 & 67.4 & 52.1 \\
ACT-SQL   & 89.5 & 85.9 & 81.8 & 63.2 & 60.9 & 42.0 & 50.0 & 31.9 & 75.0 & 60.1 \\
DAIL-SQL   & 81.9 & 67.7 & 75.3 & 57.4 & 60.9 & 37.9 & 54.8 & \textbf{46.4} & 71.2 & 54.8 \\
DCG-SQL (Ours)   & \textbf{94.8} & \textbf{93.1} & \textbf{87.0} & \textbf{78.7} & \textbf{77.0} & \textbf{57.5} & \textbf{55.4} & 43.4 & \textbf{82.1} & \textbf{72.9} \\
\bottomrule
\rowcolor{gray!25}\multicolumn{11}{c}{Deepseek-Coder-6.7B-Instruct}  \\
DIN-SQL   & 86.7 & 70.6 & 78.7 & 57.6 & 59.2 & 29.3 & 49.4 & 21.1 & 72.6 & 50.1 \\
ACT-SQL   & 88.7 & 82.7 & 81.4 & 57.4 & 66.7 & 46.6 & 41.0 & 20.5 & 74.2 & 55.7 \\
DAIL-SQL   & 85.5 & 71.0 & 79.1 & 63.0 & 64.9 & 47.7 & 48.8 & 41.6 & 73.4 & 58.9 \\
DCG-SQL (Ours)   & \textbf{91.9} & \textbf{89.9} & \textbf{86.8} & \textbf{78.0} & \textbf{74.7} & \textbf{56.3} & \textbf{56.6} & \textbf{48.8} & \textbf{81.1} & \textbf{72.5} \\
\bottomrule
\rowcolor{gray!25}\multicolumn{11}{c}{Llama 3.2-3B-Instruct}  \\
DIN-SQL   & 69.4 & 60.1 & 47.5 & 35.2 & 31.6 & 17.8 & 14.5 & 3.6 & 44.8 & 33.2 \\
ACT-SQL   & 85.9 & 77.8 & 71.7 & 56.3 & 53.4 & 37.4 & 33.7 & 24.1 & 66.0 & 53.1 \\
DAIL-SQL   & 68.1 & 50.8 & 63.7 & 49.6 & 44.8 & 25.9 & 46.4 & \textbf{42.8} & 58.8 & 44.8 \\
DCG-SQL (Ours)   & \textbf{91.9} & \textbf{87.5} & \textbf{79.8} & \textbf{71.7} & \textbf{60.9} & \textbf{48.9} & \textbf{50.6} & 41.0 & \textbf{74.7} & \textbf{66.7} \\
\bottomrule

\end{tabular}

\caption{Experimental Results on Spider Dataset with various LLMs.}
\vspace{-0.1cm}
%, comparing the execution accuracy (\%) and the exact-set-match accuracy (\%).
% The execution accuracy (EX) compares the output of the predicted SQL query to the output of the ground truth SQL query.
% The exact-set-match accuracy (EM) treats each clause as a set and compares the predicted clause to the ground truth clause. 
% Random indicates that generating SQL query with 4-shot randomly selected demonstrations.}
% \vspace{-0.1cm}
\label{main_result_1}
% \vspace{-0.1cm}
% EX indicates the correctness of the SQL query by comparing the results obtained from executing the generated query against the correct output. EM assesses whether the generated SQL query exactly matches the ground truth SQL in terms of structure and syntax. 

% \red{ACT-SQL is evaluated with a 4-shot setting due to length limitations., 이 말은 위험}}
\end{table*}

\begin{comment}
\begin{table*}[t]
%\resizebox{\linewidth}{!}
%\setlength{\tabcolsep}{0.40em}
\small
\centering
{\begin{tabular}{l|c|c|c|cc}
\toprule
\textbf{Method} & \textbf{Token Length (Avg.)} & \textbf{\# Inferences} & \textbf{Total Cost (\$)} & \textbf{EX (\%)} & \textbf{EM (\%)} \\ \toprule
Zero-shot  &  207  &  1    & 6.4  & 80.5  & 44.8         \\
Random 8-shot  &  3,082  &  1   &  95.6   & 82.1  & 60.1         \\
ACT-SQL     &  2,560  &    1    &   79.4   & 82.1  & 48.0         \\
DIN-SQL      &  2,190  &   4    &   271.7   & 82.8  & 60.1         \\
DAIL-SQL    &  743  &   1     &  23.0  & 82.1  & 71.1         \\ \midrule
\textbf{Ours}  & 1,565   &  1   & 48.5   & \textbf{83.4}  & \textbf{73.0}         \\ \bottomrule
\end{tabular}}
\caption{Experimental Results on Spider with GPT-4 (pricing: \$30 per 1M input tokens).}
%only considering input token)}}
\label{gpt_table}
\vspace{-0.2cm}
\end{table*}
\end{comment}

\section{Experiments}
\subsection{Implementation Details}
All experiments are conducted on an A100 80GB GPU. Our method requires only small size of models for efficiency. For schema link graph construction, we train the schema pruning model based on RoBERTa-large, which has approximately 355 million parameters \cite{roberta}. For schema linking, we set $\tau_{tab}$ and $\tau_{col}$ to 0.66 and 0.43, respectively. For graph-based demonstration retrieval, We train a graph encoder model based on a bi-encoder architecture~\cite{dpr}, where each encoder follows the Relation-Aware Transformer~\cite{relationtransformer} structure initialized with BERT-base~\cite{BERT}, totaling approximately 220 million parameters.
% we train a graph encoder model based on a bi-encoder architecture \cite{dpr} initialized with Bert-base \cite{bert}, which has approximately 220 million parameters in total.

For a fair comparison, we reproduce experimental results or follow existing studies. For SQL generation, we use various LLMs, including GPT-4, GPT-3.5-Turbo, DeepSeek-Coder-33B-Instruct, Llama 3.1-8B-Instruct, DeepSeek-Coder-6.7B-Instruct, and Llama 3.2-3B-Instruct~\cite{GPT-4,Deepseek,llama3}. In all experiments, the generation temperature is set to 0. We set 5 demonstrations for the few-shot setting and our input prompt is introduced in Appendix~\ref{appendix:prompt}.
% Also, we set the temperature to 0, and the input prompt we used is introduced in Appendix A.
% we set the few-shot number to 5 for our method.
% When using a small LLM as the SQL generation model, we set the few-shot number to 4 due to the limited in-context learning capabilities of sLLMs.
% When using a hyper-scaled LLM as the SQL generation model, we set the few-shot number to 8, following the existing approaches.
Training the pruning model took approximately 3 hours, while the graph encoder model was trained for 8 hours.
% pruning 모델 학습에는 약 3시간, graph encoder 모델의 경우 3번 iterate 하게 학습하였으며 약 11시간이 걸렸다. 
We used spaCy \cite{spacy} and NLTK \cite{nltk} libraries for the edges between question tokens.

\subsection{Dataset Details}
To verify the effectiveness of our method, we employ the SPIDER dataset \cite{spider}, large-scale benchmark designed for the text-to-SQL task. Spider spans 200 distinct databases across 138 domains and is widely regarded for its diverse and cross-domain nature. 
The dataset comprises 7,000 training examples and 1,034 development examples, providing a robust basis for model assessment.
The dataset categorizes SQL queries based on their complexity into four difficulty levels: easy, medium, hard, extra hard. Our evaluation experiments are conducted on the development set, using the training set as the demonstration pool for retrieval.
Moreover, using the same demonstration pool, we evaluate on Spider-DK \cite{dk}, Spider-Realistic \cite{realistic}, and Spider-Syn \cite{syn}, which are derived from Spider. They incorporate domain knowledge, remove explicitly mentioned column names, and replace schema-related words with synonyms, respectively.

\renewcommand{\arraystretch}{1.2}
\begin{table}[t]
\centering
\small
\begin{tabular}{l|cc|cc|cc}
\toprule
\multirow{2}{*}{Method} & \multicolumn{2}{c|}{DK} & \multicolumn{2}{c|}{Real} & \multicolumn{2}{c}{Syn} \\ 
\cline{2-7} \rule{0pt}{1.2EM}
 & EX & EM & EX & EM & EX & EM \\
\bottomrule
\rowcolor{gray!25}\multicolumn{7}{c}{GPT-4}  \\
ACT-SQL   & 72.0 & 47.7 & 81.3 & 55.7 & 72.1 & 48.5 \\
DAIL-SQL  & - & - & 76.0 & - & - & - \\
ASTRES    & \textbf{72.3} & 59.1 & 80.9 & 66.1 & 74.4 & 61.3 \\
DCG-SQL   & 71.6 & \textbf{66.5} & \textbf{81.9} & \textbf{73.6} & \textbf{78.7} & \textbf{68.6} \\
\bottomrule
\rowcolor{gray!25}\multicolumn{7}{c}{GPT-3.5-Turbo}  \\
DIN-SQL   & 64.3 & 36.6 & 65.2 & 41.1 & 65.3 & 34.7 \\
ACT-SQL   & 68.2 & 45.4 & 76.4 & 48.6 & 70.4 & 44.4 \\ 
DAIL-SQL  & - & - & 67.9 & - & - & - \\
ASTRES    & 68.8 & 49.3 & 78.0 & 60.8 & 66.9 & 51.2 \\
DCG-SQL   & \textbf{69.0} & \textbf{59.4} & \textbf{79.0} & \textbf{62.8} & \textbf{74.2} & \textbf{61.6} \\
\bottomrule
\rowcolor{gray!25}\multicolumn{7}{c}{Deepseek-Coder-33B-Instruct}  \\
% DIN-SQL   & \hl{00.0} & \hl{00.0} & \hl{00.0} & \hl{00.0} & \hl{00.0} & \hl{00.0} \\
ACT-SQL   & 69.5 & 47.5 & 76.4 & 51.8 & 69.1 & 49.3 \\
ASTRES    & \textbf{70.5} & 46.4 & 77.4 & 59.3 & 68.7 & 49.5 \\
DCG-SQL   & 68.4 & \textbf{62.8} & \textbf{79.1} & \textbf{71.1} & \textbf{73.6} & \textbf{62.6} \\
\bottomrule
\rowcolor{gray!25}\multicolumn{7}{c}{Llama-3.1-8B-Instruct}  \\
DIN-SQL   & 56.3 & 44.1 & 59.6 & 36.4 & 57.2 & 43.5 \\
ACT-SQL   & 63.6 & 48.0 & 74.4 & 60.2 & 66.6 & 53.7 \\
DCG-SQL   & \textbf{66.5} & \textbf{60.2} & \textbf{75.8} & \textbf{67.1} & \textbf{70.1} & \textbf{59.7} \\
\bottomrule
\rowcolor{gray!25}\multicolumn{7}{c}{Deepseek-Coder-6.7B-Instruct}  \\
DIN-SQL   & 54.1 & 16.2 & 53.5 & 29.3 & 53.7 & 25.4 \\
ACT-SQL   & 63.2 & 44.9 & 69.1 & 53.5 & 63.2 & 44.6 \\
DCG-SQL   & \textbf{65.0} & \textbf{55.5} & \textbf{74.8} & \textbf{68.5} & \textbf{70.5} & \textbf{59.9} \\
\bottomrule
\rowcolor{gray!25}\multicolumn{7}{c}{Llama-3.2-3B-Instruct}  \\
DIN-SQL   & 35.7 & 27.5 & 38.6 & 29.1 & 39.0 & 27.2 \\
ACT-SQL   & 54.4 & 41.7 & 62.0 & 47.0 & 57.4 & 44.4 \\
DCG-SQL   & \textbf{59.1} & \textbf{52.5} & \textbf{68.5} & \textbf{59.6} & \textbf{60.9} & \textbf{51.9} \\

% % ACT-SQL   & 92.3 & 85.9 & 89.3 & 59.1 & 80.2 & 53.1 & 61.5 & 18.9 & 83.9 & 57.9 \\
% % DAIL-SQL   & 91.1 & 90.3 & 88.6 & 76.0 & 75.9 & 60.3 & 62.0 & 50.0 & 82.8 & 72.6 \\
% % ASTRES (w/ Graphix-T5)&-&- & -& -& - & - & - &-& 86.6 &77.3 \\
% % DCG-SQL   & \textbf{95.6} & \textbf{92.3} & \textbf{90.6} & \textbf{83.0} & \textbf{82.8} & \textbf{70.7} & \textbf{72.3} & \textbf{61.4} & \textbf{87.5} & \textbf{79.7} \\
% (1) & 77.0  & 66.2         \\
% (2) & 77.1  & 65.6        \\
% (3) & 76.2 & 66.5         \\ \midrule
% Ours            & \textbf{80.1}  & \textbf{71.8}         \\
\bottomrule
\end{tabular}
\caption{Experimental Results on Spider-DK, Spider-Realistic, and Spider-Syn with various LLMs.}
\label{main_result_2}
% \vspace{-0.3cm}
\end{table}

\subsection{Baselines \& Evaluation Metrics}

We compared our method with recent state-of-the-art in-context learning baselines for Text-to-SQL~\cite{DIN, ACT, DAIL, ast}.
% For a fair comparison, we compare recent state-of-the-art methods~\cite{DIN, ACT, DAIL} that utilize large LLMs for Text-to-SQL and provide publicly available code.
% Due to the lack of publicly available code, some approaches \cite{Enhancing, MCS, purple} could not be compared.}
% \red{Other approaches proposed in \cite{Enhancing, ast, MCS, purple} were not compared, as their code or prompts are not publicly available.} 
%%% traditional 한것들과 비교하지 않겠따는 말
% Also, since our focus is improving in-context learning for text-to-SQL, we do not include comparisons with training model-based approaches \cite{graphix, RESDSQL, RASAT}.
%%% traditional 한것들과 비교하지 않겠따는 말
% Also, since our focus is improving text-to-SQL performance of in-context learning, we did not compare with training model-based approaches \cite{graphix, RESDSQL, RASAT}.
% \red{The approaches are evaluated across five different LLMs of different sizes, from the largest to the smallest: GPT-4, GPT-3.5-Turbo, Deepseek-Coder-33B-Instruct~\cite{deepseek}, Llama 3.1 8B-Instruct~\cite{llama3} and Deepseek-Coder-6.7B-Instruct \cite{deepseek}}.

% The execution accuracy (EX) compares the output of the predicted SQL query to the output of the ground truth SQL query.
% The exact-set-match accuracy (EM) treats each clause as a set and compares the predicted clause to the ground truth clause. Random indicates that generating SQL query with 4-shot randomly selected demonstrations.

We use two evaluation metrics, Execution Accuracy (EX) and Exact-set-match Accuracy (EM). Execution accuracy measures whether the predicted SQL query yields the same result as the ground truth query when executed.
% The execution accuracy compares the output of the predicted SQL query to the output of the ground truth SQL query.
Exact-set-match accuracy treats each clause as a set and compares the predicted clause to the ground truth clause.
% \footnote{\red{All evaluation metrics are computed strictly following https://github.com/taoyds/test-suite-sql-eval}}.

\subsection{Experimental Results}

Table \ref{main_result_1} presents the Text-to-SQL performance on the Spider dataset using various hyper-scaled LLMs and small LLMs.
Our proposed method consistently achieves higher execution accuracy and exact match accuracy compared to existing approaches.
% For hyper-scaled LLMs, ou
In particular, our method achieves notable improvements for sLLMs. For execution accuracy (EX), it increases by 9.5\% with Llama 3.1-8B and 9.3\% with DeepSeek-Coder-6.7B. Notably, there is a substantial gap in exact match accuracy (EM), increasing by 21.3\% with Llama 3.1-8B and 23.1\% with DeepSeek-Coder-6.7B. This suggests that our demonstrations provide the precise patterns needed for the test cases, as in-context learning relies on pattern following~\cite{gpt3, in-context}.
These results indicate that our approach, which leverages deep contextual schema link graph-based retrieval for SQL generation, effectively utilizes the in-context learning capabilities, leading to outperforming existing methods in text-to-SQL task.

Additionally, to demonstrate the effectiveness of our approach in more challenging scenarios, we conduct experiments on Spider-DK, Spider-Realistic, and Spider-Syn for SQL generation, as shown in Table \ref{main_result_2}.
Our method consistently demonstrates strong performance across various LLM scenarios and datasets. These findings highlight the effectiveness of our method across different datasets, showing strong performance not only with hyper-scaled LLMs but also with small LLMs.

\subsection{Ablation Studies}
%%%%%%%%%%%%%%%%% JSL %%%%%%%%%%%%%%%
\begin{comment}
\paragraph{Effectiveness of our proposed modules.}
Table~\ref{ablation_1} presents the performance of our method on the Spider dataset with different combinations of our proposed modules.
In case (1), it uses only 4-shot randomly selected demonstrations, and the DB schema prompt for generating the SQL query is not pruned.
In case (2), it uses 4-shot demonstrations based on our retrieval method, and the DB schema prompt is not pruned.
Case (3) uses 4-shot random demonstrations, and the DB schema prompt is pruned based on our SQL generation method.
In our case, both our retrieval and SQL generation methods are applied.

When our method is not applied, it achieves 71.8\% and 51.2\% for EX and EM accuracy, respectively. However, when both GDR and GPD are utilized, the model's accuracy significantly increases to 80.1\% and 71.8\%, compared to using each module individually.
\end{comment}
%%%%%%%%%%%%%%%%% JSL %%%%%%%%%%%%%%%

\renewcommand{\arraystretch}{1.2}
\begin{table}[t]
\centering
\small
\begin{tabular}{l|cc|cc|cc}
\toprule
\multirow{2}{*}{Retrieval} & \multicolumn{2}{c|}{Spider} & \multicolumn{2}{c|}{Real}& \multicolumn{2}{c}{Syn} \\ 
\cline{2-7} \rule{0pt}{1.2EM}
 & EX & EM & EX & EM & EX & EM \\
\bottomrule
\rowcolor{gray!25}\multicolumn{7}{c}{GPT-3.5-Turbo}  \\
Random    & 78.8 & 47.0 & 74.6 & 42.2 & 60.0 & 38.4 \\
DCG-SQL   & \textbf{83.3} & \textbf{70.7} & \textbf{79.0} & \textbf{62.8} & \textbf{74.2} & \textbf{61.6} \\
\bottomrule
\rowcolor{gray!25}\multicolumn{7}{c}{Llama-3.1-8B-Instruct}  \\
Random    & 73.0 & 48.5 & 69.7 & 45.5 & 61.6 & 39.7 \\
DCG-SQL   & \textbf{82.1} & \textbf{72.9} & \textbf{75.8} & \textbf{67.1} & \textbf{70.1} & \textbf{59.7} \\
\bottomrule
\end{tabular}
\caption{Effectiveness of our retrieved demonstrations.}
\label{abla_random}
% \vspace{-0.3cm}
\end{table}

\begin{table}[t]
\centering
\small
\begin{tabular}{c|cc}
\toprule
Retrieval Case & EX (\%) & EM (\%) \\ \toprule
(1) & 77.0  & 66.2        \\
(2) & 77.1  & 65.6        \\
(3) & 76.2 & 66.5         \\ 
(4) & 76.1 & 59.6       \\ \midrule
Ours            & \textbf{82.1}  & \textbf{72.9}         \\
\bottomrule
\end{tabular}
\caption{Effectiveness of our deep contextual schema link graph-based retrieval. The experiments were conducted on Spider with Llama-3.1-8B-Instruct.}
\label{abla_retrieval}
\vspace{0.2cm}
\end{table}
%순서대로 masked, 

\begin{table}[t]
\centering
\small
\begin{tabular}{l|ccc} 
\toprule
DB Schema Type  & Precision & Recall & F1 Score\\ \midrule
Table & 96.2 & 97.4 & 96.8 \\ 
Column & 90.7 & 95.5 & 93.1 \\ \bottomrule
\end{tabular}
\caption{Schema pruning performance.}
\label{abla_pruning}
% \vspace{-0.1cm}
\end{table}

\begin{table}[t]
%\resizebox{\linewidth}{!}
\small
\centering
{\begin{tabular}{l|cc}
\toprule
Method & EX (\%) & EM (\%) \\ \toprule
DAIL-SQL            & 71.2 & 54.8         \\
DAIL-SQL w/ Our Retrieval   &  \textbf{75.6} & \textbf{61.3}         \\
\bottomrule
\end{tabular}}
\caption{Transferability of our proposed modules. The experiments were conducted on Spider with Llama-8B.}
\label{abla_transfer}
% \vspace{-0.3cm}
\end{table}

\begin{table}[t]
\centering
\small
\begin{tabular}{c|c}
\toprule
Method & Average Tree Edit Distance \\ \toprule
ACT-SQL & 27.17         \\
DAIL-SQL & 22.38        \\
% Ours & \textbf{68.74}   \\
DCG-SQL & \textbf{13.82}   \\
\bottomrule
\end{tabular}
\caption{Average Tree Edit Distance between demonstration and target SQL query on Spider Dataset}
\label{abla_tree}
\vspace{-0.2cm}
\end{table}

\paragraph{Effectiveness of our retrieval method.}
Table \ref{abla_random} presents the results when demonstrations are randomly chosen.
Existing approaches \cite{ACT, DAIL} failed to maximize in-context learning capabilities due to their ineffective retrieval approaches as shown in Table \ref{intro_table}, whereas our method consistently achieves performance improvements across various datasets and LLMs.

To demonstrate the effectiveness of our retrieval method, we compare it with various retrieval approaches, as shown in Table \ref{abla_retrieval}.
For demonstration retrieval, case (1) is based on the text embedding of question and (2) is based on the masked question text embedding~\cite{GPT-3.5}. Case (3) is based on both question and DB schema, but uses text embeddings. In case (4), it is based on the existing graph representation \cite{graphix}. Cases (1)–(3) use BERT-base as the text encoder, whereas case (4) employs the same graph encoder as our method. All (1)–(4) are trained using the same self-supervised learning approach described in Section~\ref{sec:3.2}
% To demonstrate the effectiveness of our retrieval method, we compare with various retrieval approaches as shown in Table \ref{ablation_2}. For demonstrations retrieval, case (1) and (2) is based on question text embedding and masked question text embedding, respectively. In case (3), it is based on both question and Db schema, but it is a text embedding. In our case, it is based on both question and DB schema, and it is a graph embedding.

As shown in cases (1) and (2), demonstration retrieval based only on question does not achieve good performance. Also, retrieval methods that consider both question and DB schema information without their contextual relationships, show poor performance. In contrast, our method captures contextual relationships in the joint representation of Text-to-SQL samples, leading to better demonstration retrieval.

\paragraph{Effectiveness of our DB schema pruning.}
We demonstrate the effectiveness of our pruning method as shown in Table \ref{abla_pruning}. For tables, our method achieves a precision of 96.2\%, recall of 97.4\%, and an F1 score of 96.8\%. For columns, our method shows a precision of 90.7\%, recall of 95.5\%, and an F1 score of 93.1\%. These results show that our method effectively identifies relevant schema items and exclude irrelevant links.
% demonstration retrieval and SQL generation.
% 우리는 pruning 모델이 유효한 schema를 잘 identify하는지 평가하였고, 이러한 결과를 \cite{RESDSQL}의 ranking enhanced encoder 모델의 성능과 비교하였다. Table \ref{ablation_4}는 이 결과를 나타낸다. our pruning 모델을 사용하였을때, table과 column 각각 96.8\%, 93.0\%의 준수한 분류 성능을 보였으며, 이는 \cite{RESDSQL}의  유사한 결과를 보였다. 이는 우리 pruning 모델의 성능이 준수함을 나타내며, 이 모델을 인퍼런스를 통해 스키마 링킹을 동시에 진행할 수 있기 때문에, pruning 모델이  graph construction에 효과적임을 나타낸다.

\paragraph{Transferability of our proposed modules.}
% To demonstrate transferability of our proposed modules, we applied our modules to DAIL-SQL approach~\cite{DAIL}. In Table 6, all experiment results는 DAIL-SQL의 prompt templete을 사용한 결과이다. 우리의 방법으로 retrieval한 demonstration sets을 사용하면서, prompt templete은 DAIL-SQL을 따랐을 때의 성능은 EX에 대해서는 72.8\%과 EM에 대해서는 64.2\% 였다, respectively. 우리의 방법으로 unrelevant schema를 exclude한 방법을 적용하면서, prompt templete은 DAIL-SQL을 따랐을 때의 성능은 EX에 대해서는 74.0\%과 EM에 대해서는 58.7\% 였다, respectively. 우리의 방법을 DAIL-SQL에 둘다 적용했을 때의 성능은 75.5\%와 68.8\%으로, original DAIL-SQL 보다 6.0\%와 28.1\%의 성능 향상을 보인다. 이러한 결과는 우리의 방법이 다양한 prompt templete에 대해서도 효과적으로 적용될 수 있음을 보여준다.
To demonstrate the transferability of our modules, we applied our retrieval to the DAIL-SQL~\cite{DAIL}. All the experimental results in Table \ref{abla_transfer} are obtained using the same setting as DAIL-SQL. When we use the demonstrations sets retrieved by our method, the performance is 75.6\% in EX and 61.3\% in EM, respectively. 
% Additionally, when we apply our method to exclude irrelevant schemas while using the DAIL-SQL prompt template, the performance is 73.3\% in EX and 54.8\% in EM. When both of our methods are applied to DAIL-SQL, the performance improves to 73.8\% in EX and 65.5\% in EM, 
This corresponds to improvements of 6.2\% and 11.9\% over the original DAIL-SQL. These results demonstrate that our retrieval method can be effectively applied to various prompt templates and Text-to-SQL methods.

\begin{comment}
\paragraph{Ablation study for few-shot demonstrations.}
To demonstrate the effectiveness of our method across various few-shot numbers, we conduct experiments with various few-shot settings, \red{including zero-shot}. As shown in Figure \ref{fewshot}, our method outperforms existing methods not only with the 4-shot setting used in the main results, but across \red{all numbers of shots}. This indicates that our method can effectively generate target SQL queries. 
% our proposed method가 various few-shot number에 대해서 효과적임을 보이기 위해서, 우리는 few-shot 개수에 따른 실험들을 수행하였다.
% As shown in Figure \ref{fewshot}, 우리의 방법은 main results에서 사용했던 8개에서 뿐만 아니라, 모든 few-shot number에 대해서 기존 방법들보다 좋은 성능을 보인다.
% 이는 우리의 방법이 small LLM만을 사용하고도 target SQL query를 효과적으로 generation 한다는 것을 보여준다.

% In the main results, we set the number of few-shot demonstrations to 8. 
\end{comment}

\paragraph{Analysis of retrieved demonstrations.}
To analyze the quality of retrieved demonstrations, we calculate the average Tree Edit Distance (TED) \cite{syntaxsqlnet}, which measures the similarity between the target SQL of user input and the SQLs of the retrieved demonstrations. 
If TED is small, it indicates that the SQLs of retrieved demonstrations are similar to the target SQL of user input. As shown in the Table \ref{abla_tree}, our method achieves the lowest TED compared to existing approaches. Since in-context learning of LLMs tends to follow the patterns in the demonstrations \cite{gpt3, in-context}, the result supports that our proposed method effectively retrieves useful demonstrations for generating correct SQL.

% \begin{table}[t]
% \centering
% \small
% \begin{tabular}{c|c}
% \toprule
% Method & Average Tree Edit Distance \\ \toprule
% ACT-SQL & 27.17         \\
% DAIL-SQL & 22.38        \\
% % Ours & \textbf{68.74}   \\
% DCG-SQL & \textbf{13.82}   \\
% \bottomrule
% \end{tabular}
% \caption{Average Tree Edit Distance between demonstration and target SQL query on Spider Dataset}
% \label{abla_tree}
% % \vspace{-0.3cm}
% \end{table}

\begin{table}[t]
\centering

\small
\begin{tabular}{l|cccc} 
\toprule
Method & EX (\%) & \makecell{Retrieval \\ Params.} & \makecell{LLM \\ Calls} & \makecell{Retrieval\\Latency }\\ 
\midrule
DIN-SQL & 67.4 & - & 4 & -\\ 
ACT-SQL & 75.0 & 1M & 1 & 0.7\\ 
DAIL-SQL & 71.2 & 3B & 1 & 9.1\\ 
DCG-SQL & 82.1 & 0.5B & 1 & 1.3\\ 
\bottomrule
\end{tabular}

\caption{Comparison of Efficiency. Experimental results are measured on Spider with Llama-3.1-8B-Instruct.}
\label{efficiency table}
% \vspace{-0.1cm}
\end{table}

\paragraph{Efficiency of our retrieval method.}
To demonstrate the efficiency of our method, we evaluate the inference latency on the Spider dataset, as shown in Table~\ref{efficiency table}. Retrieval latency refers to the time required on average to extract embeddings from test cases. DIN-SQL~\cite{DIN}, which performs four separate LLM inferences without retrieval, incurs four times the LLM latency compared to other methods. Meanwhile, DAIL-SQL~\cite{DAIL} employs a retrieval model with 3B parameters~\cite{graphix}, resulting in latency up to 9.1 seconds. Although ACT-SQL~\cite{ACT} shows relatively low inference latency, its execution accuracy remains unsatisfactory. In contrast, our method requires only a single LLM inference and employs lightweight retrieval modules with just 0.5B trainable parameters, including the schema pruning model and the graph encoder. This design enables more efficient retrieval while maintaining high execution accuracy. These results clearly demonstrate the effectiveness and efficiency of our approach.

%% file: texs/5.conclusion.tex
\section{Conclusion}
This paper presents a novel approach for retrieving demonstrations and generating SQL queries, effective across LLMs of various sizes.
Our method leverages a deep contextual schema link graph to enhance demonstration retrieval and SQL generation. It achieved strong performance across hyper-scaled and small LLMs.
% , including GPT-4, GPT-3.5-Turbo, DeepSeek-Coder-33B, Llama 3.1-8B, and DeepSeek-Coder-6.7B.}

\section*{Limitations}
Under the same LLM models, our method outperforms the existing SOTA. Our method has shown highly effective in improving performance with small LLMs, however,  
there is still a performance gap between the use of hyper-scaled LLMs and small LLMs. Future work should focus on addressing this performance gap effectively.

\section*{Acknowledgments}
This work was partly supported by Institute of Information \& communications Technology Planning \& Evaluation (IITP) grant funded by the Korea government(MSIT) (No.IITP-2025-RS-2025-00437633, Information Technology Research Center, 20\%), (No.RS-2019-II190421, AI Graduate School Support Program(Sungkyunkwan University), 20\%), (No.RS-2025-02218768, Accelerated Insight Reasoning via Continual Learning, 20\%). This work was also supported by the National Research Foundation of Korea(NRF) grant funded by the Korea government(MSIT) (No.RS-2025-00521391, 20\%) and Artificial intelligence industrial convergence cluster development project funded by the Ministry of Science and ICT(MSIT, Korea)\&Gwangju Metropolitan City (20\%).

%% file: texs/6.appendix.tex
\clearpage
\appendix

\begin{figure*}[t!]
  \includegraphics[width=\linewidth]{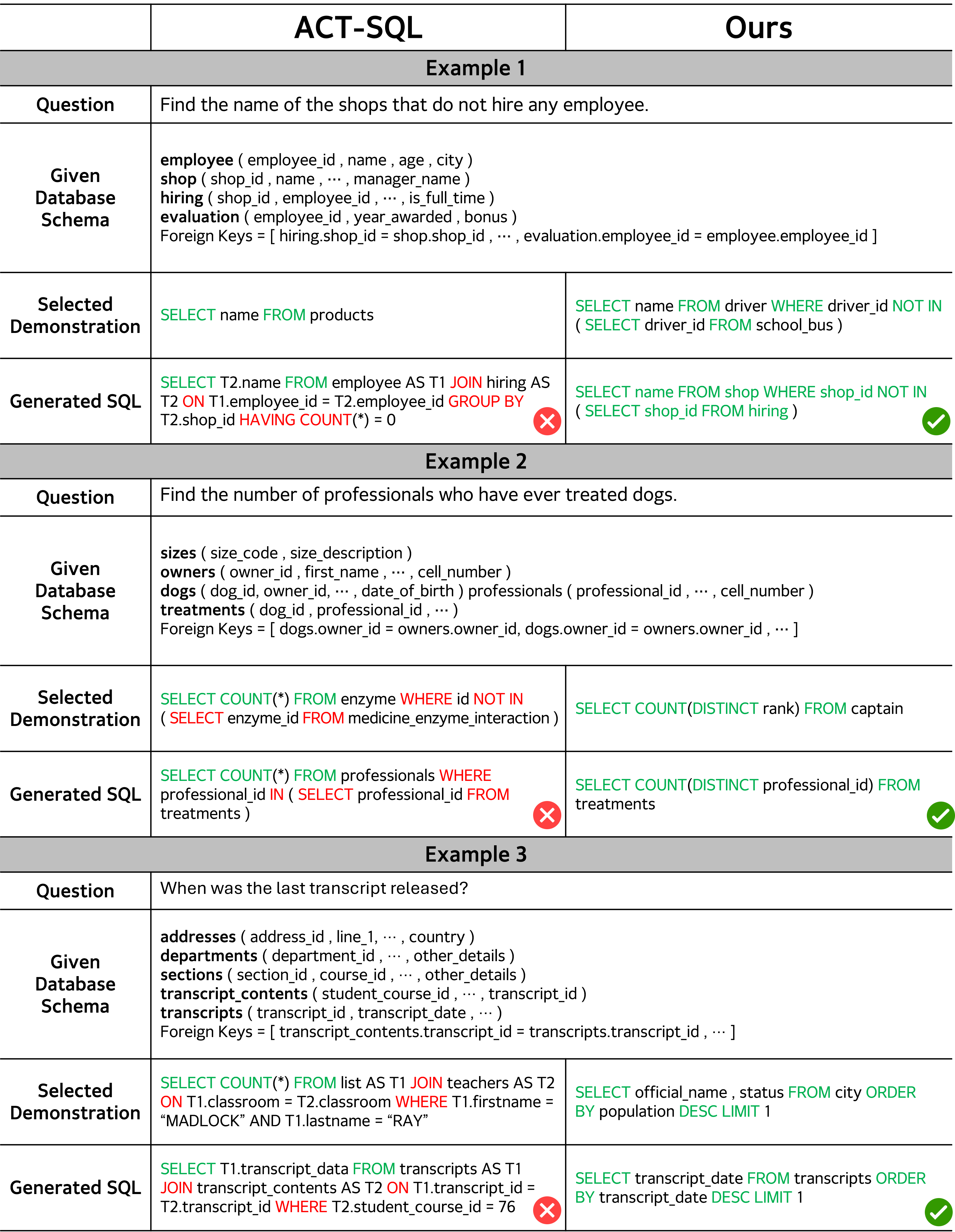} 
  \caption {Qualitative results. These results demonstrate a comparison between our generated queries and those from ACT-SQL.}
   
    \label{dail qualitative results}
% \vspace{-0.5cm}
\end{figure*}

\begin{figure*}[t!]
  \includegraphics[width=\linewidth]{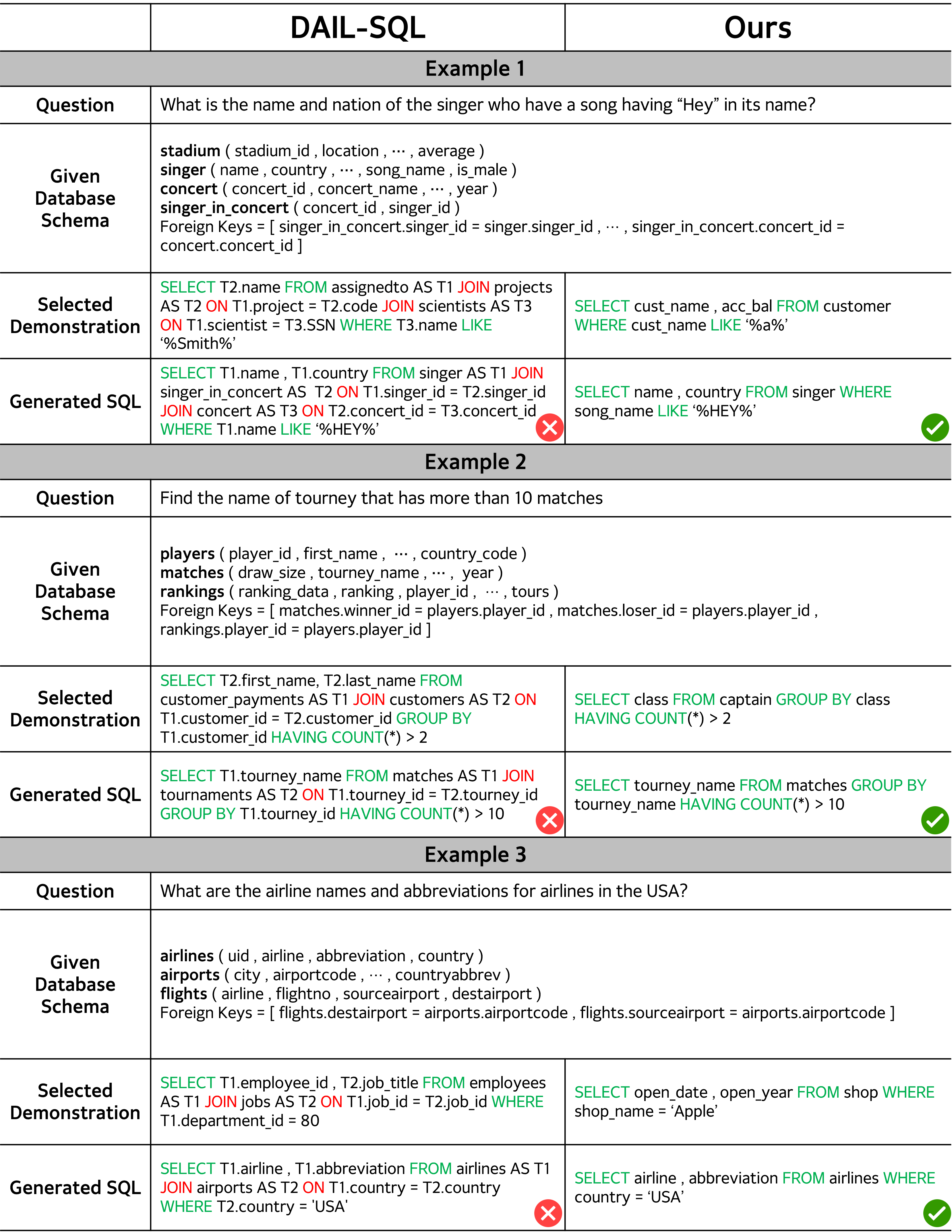} 
  \caption {Qualitative results. These results demonstrate a comparison between our generated queries and those from DAIL-SQL.}
   
    \label{act qualitative results}
% \vspace{-0.5cm}
\end{figure*}

\section{Generalization Performance on BIRD dataset}
To verify the generalization ability of our method, we conduct experiments on the BIRD dataset using the same hyperparameters and evaluate its performance on the BIRD dev dataset. ACT-SQL~\cite{ACT} cannot be evaluated on the BIRD dataset because it relies heavily on the Spider dataset. In addition, DIN-SQL~\cite{DIN} cannot be evaluated with Llama 3.1-8B-Instruct model because input prompts, provided in their official code, exceed the maximum token length of the model which is 8,192. As shown in the Table \ref{ablation_bird}, existing methods exhibit similar performance to random demonstrations, while our method outperforms them, indicating its strong generalization performance.
\begin{table}[htbp]
\centering
\small
\begin{tabular}{l|cc}
\toprule
Method & EA & VES \\ \bottomrule
\rowcolor{gray!25}\multicolumn{3}{c}{GPT-3.5-Turbo}  \\
DIN-SQL & 45.1 & 52.4 \\
DAIL-SQL & 45.9 & 50.6 \\
DCG-SQL & \textbf{49.7} & \textbf{54.4} \\ 
\bottomrule
\rowcolor{gray!25}\multicolumn{3}{c}{Deepseek-Coder-33B-Instruct}  \\
DIN-SQL & 48.8 & 53.4 \\
DAIL-SQL & 49.9 & 52.6 \\
DCG-SQL & \textbf{53.2} & \textbf{56.0} \\
\bottomrule
\rowcolor{gray!25}\multicolumn{3}{c}{Llama 3.1-8B-Instruct}  \\
DAIL-SQL & 38.40  & 39.38        \\
DCG-SQL & \textbf{42.20} & \textbf{46.67}        \\
\bottomrule
\rowcolor{gray!25}\multicolumn{3}{c}{Deepseek-Coder-6.7B-Instruct}  \\
DIN-SQL & 41.7 & 44.8 \\
DAIL-SQL & 40.7 & 43.8 \\
DCG-SQL & \textbf{44.8} & \textbf{47.2} \\
\bottomrule
\end{tabular}
\caption{Performance on BIRD dataset with various LLMs. EA and VES indicate Execution Accuracy and Valid Efficiency Score, respectively.}
\label{ablation_bird}
% \vspace{-0.3cm}
\end{table}

\section{Analysis of Qualitative Results}
To ensure a reliable qualitative analysis, we compare the generated SQL queries with those from existing approaches \cite{DAIL, ACT} by using diverse cases of questions and DB schemas.
Figure \ref{dail qualitative results} and Figure \ref{act qualitative results} demonstrate the question, database, and selected demonstrations, which are used for SQL query generation.
In the existing approaches, the given database contains all schema items, making the model choose irrelevant schema items for SQL query generation. Additionally, the selected demonstrations differ significantly in syntax from the target SQL query. Such difference cause the model to generate incorrect SQL queries since in-context learning makes LLMs follow the patterns in the examples~\cite{gpt3}. In contrast, our method effectively includes relevant schema items and the selected demonstrations contain the highly similar to target SQL query. These indicate that our method successfully identifies the useful DB schema information and demonstrations for generating the target SQL.

% \section{Experimental Results using sLLM with 8-shots}
% To verify that our proposed method using sLLM is effective across various few-shot settings, we conducted experiments as shown in Table~\ref{8shot}. The results show that our method outperforms existing approaches in the 8-shots.
% To verify that our propose method using sLLM이 various한 few-shots 환경에서도 효과적임을 verify 하기 위해, We conduct the experiments as shown in Table \ref{tableB}.
% It shows that our method outperforms existing approaches with 8 shot setting.

\section{Prompt Details}
\label{sec:appendix}
In this section, we presents all the prompts of our methods. Our prompt is based on API DOCS format \cite{EvaluateText-to-SQL}. Following \cite{bridge}, we also provide potentially useful values by performing if there is match between question tokens and the values of each column.

\subsection{Test Case Prompt}
\label{appendix:prompt}
We demonstrate the prompt for a test case in Table~\ref{testcase_prompt}.
% \noindent \textbf{Table city}, columns = [*, City\_ID, Official\_Name, Status, Area\_km\_2, Population, Census\_Ranking] \\

\subsection{Automated-CoT prompt}
\label{appendix:automated-cot}
As shown in Table \ref{cot_prompt}, we categorized the training samples based on the SQL into four cases, SIMPLE, JOIN, NESTED, and IUEN. We designed automated-CoT prompt beginning with "Let's think step by step." except for SIMPLE cases. For JOIN, we make the model to generate intermediate step first, then generate final query~\cite{DIN}. For nested and IEUN(INTERSECT, EXCEPT, UNION, NOT IN), we added a natural language phrase to describe briefly about the nested SQL query by converting WHERE clause and HAVING clause of the nested query with rule-based approach \cite{editable}.

\begin{table*}[htbp]
%\resizebox{\linewidth}{!}
\setlength{\tabcolsep}{0.40em}
\small
\centering
{\begin{tabular}{l}
\toprule
 Prompt  \\ \toprule
\\
\#\#\# Answer the question by SQLite SQL query only and with no explanation.\\
\#\#\# SQLite SQL tables, with their properties: \\
\# \\
\# city (id, name, countrycode, district, population) \\
\# sqlite\_sequence (name,seq) \\
 \# country (code, name, continent, region, surfacearea, indepyear, population, \\
 lifeexpectancy, gnp, gnpold, localname, governmentform, headofstate, capital, code2)\\
 \# countrylanguage (countrycode, language [Dutch, English], isofficial, percentage) \\
 \# Foreign Keys = [city.countrycode = country.code, countrylanguage.countrycode = country.code] \\
 \# \\
 \#\#\# Question: Which regions speak Dutch or English?\\

 % \#\#\# Answer the question by SQLite SQL query only and with no explanation. \\
 % \#\#\# SQLite SQL tables, with their properties: \\
 % \# \\
 % \# country (code, region) \\
 % \# countrylanguage (countrycode, language [Dutch, English]) \\
 % \# Foreign Keys = [city.countrycode = country.code, countrylanguage.countrycode = country.code] \\
 % \# \\
 % \#\#\# Question: Which regions speak Dutch or English? \\

\\

 \bottomrule
\end{tabular}}
\caption{Test Case Prompt Examples.}
\label{testcase_prompt}
\end{table*}   

\begin{table*}[t!]
% \resizebox{\linewidth}{!}
\setlength{\tabcolsep}{0.40em}
% \small
\scriptsize
\centering
{\begin{tabular}{l|l}
\toprule
 & Prompt  \\ \toprule
Simple
& \\
&\#\#\# Answer the question by SQLite SQL query only and with no explanation.\\
&\#\#\# SQLite SQL tables, with their properties: \\
&\# \\
&\# head (age,...) \\
& \# Foreign Keys = [...] \\
& \# \\
& \#\#\# Question: How many heads of the departments are older than 56 ?\\
& \#\#\# SQL: SELECT COUNT(*) FROM head WHERE age > 56 ;\\

\\ \midrule
JOIN & \\
& \#\#\# Answer the question by SQLite SQL query only and with no explanation. \\
& \#\#\# SQLite SQL tables, with their properties: \\
& \# \\
& \# station (id, name, long, ...) \\
& \# trip (duration, start\_station\_id, ...) \\
& \# Foreign Keys = [...] \\
& \# \\
& \#\#\# Question: For each station, return its longitude and the average duration of trips \\
& that started from the station. \\
& Let`s think step by step. we need to join the tables 'station' and 'trip'. \\
& Create an intermediate representation, then use it to construct the query. \\
& Intermediate representation: "FROM station AS T1 JOIN trip AS T2 ON T1.id = T2.start\_station\_id". \\
& \#\#\# SQL: SELECT T1.name , T1.long , AVG(T2.duration) FROM station AS T1 JOIN trip AS T2 \\
& ON T1.id = T2.start\_station\_id GROUP BY T2.start\_station\_id ; \\

\\ \midrule
NESTED & \\
& \#\#\# Answer the question by SQLite SQL query only and with no explanation. \\
& \#\#\# SQLite SQL tables, with their properties: \\
& \# \\
& \# weather (date, min\_dew\_point\_f, zip\_code, ....) \\
& \# Foreign Keys = [...] \\
& \# \\
& \#\#\# Question: On which day and in which zip code was the min dew point lower than \\
& any day in zip code 94107? \\
& Let`s think step by step. we need a nested subquery for 'zip code of the weather is 94107'. \\
& Nested subquery: "( SELECT min ( min\_dew\_point\_f ) FROM weather WHERE zip\_code = 94107 )". \\
& With the nested subquery, we can get the final SQL query. \\
& \#\#\# SQL: SELECT date , zip\_code FROM weather \\
& WHERE min\_dew\_point\_f < (SELECT MIN(min\_dew\_point\_f) FROM weather \\
& WHERE zip\_code = 94107) ; \\

\\ \midrule
IUEN & \\
& \#\#\# Answer the question by SQLite SQL query only and with no explanation. \\
& \#\#\# SQLite SQL tables, with their properties: \\
& \# \\
& \# city (status, population,...) \\
& \# Foreign Keys = [...] \\
& \# \\
& \#\#\# Question: Show the status shared by cities with population bigger than 1500 \\
& and smaller than 500. \\
& Let`s think step by step. The question can be solved using the 'INTERSECT' set operator \\
& and two subqueries: one for 'population of the city is above 1500' and the other for  \\
& 'population of the city is less than 500'. \\
& First subquery: SELECT Status FROM city WHERE Population > 1500 \\
& Second subquery: SELECT Status FROM city WHERE Population < 500 \\
& With the nested subquery, we can get the final SQL query. \\
& \#\#\# SQL: SELECT status FROM city WHERE population > 1500 INTERSECT \\
& SELECT status FROM city WHERE population < 500 ; \\

\\ \bottomrule
\end{tabular}}
\caption{Automated-CoT for demonstrations.}
\label{cot_prompt}
\end{table*}

%% file: main.bbl
\begin{thebibliography}{39}
\providecommand{\natexlab}[1]{#1}

\bibitem[{Bird and Loper(2004)}]{nltk}
Steven Bird and Edward Loper. 2004.
\newblock \href {https://aclanthology.org/P04-3031} {{NLTK}: The natural language toolkit}.
\newblock In \emph{Proceedings of the {ACL} Interactive Poster and Demonstration Sessions}, pages 214--217, Barcelona, Spain. Association for Computational Linguistics.

\bibitem[{Brown et~al.(2020)Brown, Mann, Ryder, Subbiah, Kaplan, Dhariwal, Neelakantan, Shyam, Sastry, Askell et~al.}]{gpt3}
Tom Brown, Benjamin Mann, Nick Ryder, Melanie Subbiah, Jared~D Kaplan, Prafulla Dhariwal, Arvind Neelakantan, Pranav Shyam, Girish Sastry, Amanda Askell, et~al. 2020.
\newblock Language models are few-shot learners.
\newblock \emph{Advances in neural information processing systems}, 33:1877--1901.

\bibitem[{Chang and Fosler-Lussier(2023)}]{howtoprompt}
Shuaichen Chang and Eric Fosler-Lussier. 2023.
\newblock How to prompt llms for text-to-sql: A study in zero-shot, single-domain, and cross-domain settings.
\newblock \emph{arXiv preprint arXiv:2305.11853}.

\bibitem[{DeepSeek-AI et~al.(2024)DeepSeek-AI, :, Bi, Chen, Chen, Chen, Dai, Deng, Ding, Dong, Du, Fu, Gao et~al.}]{Deepseek}
DeepSeek-AI, :, Xiao Bi, Deli Chen, Guanting Chen, Shanhuang Chen, Damai Dai, Chengqi Deng, Honghui Ding, Kai Dong, Qiushi Du, Zhe Fu, Huazuo Gao, et~al. 2024.
\newblock \href {https://arxiv.org/abs/2401.02954} {Deepseek llm: Scaling open-source language models with longtermism}.
\newblock \emph{Preprint}, arXiv:2401.02954.

\bibitem[{Deng et~al.(2021)Deng, Awadallah, Meek, Polozov, Sun, and Richardson}]{realistic}
Xiang Deng, Ahmed~Hassan Awadallah, Christopher Meek, Oleksandr Polozov, Huan Sun, and Matthew Richardson. 2021.
\newblock \href {https://doi.org/10.18653/v1/2021.naacl-main.105} {Structure-grounded pretraining for text-to-{SQL}}.
\newblock In \emph{Proceedings of the 2021 Conference of the North American Chapter of the Association for Computational Linguistics: Human Language Technologies}, pages 1337--1350, Online. Association for Computational Linguistics.

\bibitem[{Devlin et~al.(2019)Devlin, Chang, Lee, and Toutanova}]{BERT}
Jacob Devlin, Ming-Wei Chang, Kenton Lee, and Kristina Toutanova. 2019.
\newblock \href {https://doi.org/10.18653/v1/N19-1423} {{BERT}: Pre-training of deep bidirectional transformers for language understanding}.
\newblock In \emph{Proceedings of the 2019 Conference of the North {A}merican Chapter of the Association for Computational Linguistics: Human Language Technologies, Volume 1 (Long and Short Papers)}, pages 4171--4186, Minneapolis, Minnesota. Association for Computational Linguistics.

\bibitem[{Dubey et~al.(2024)Dubey, Jauhri, Pandey, Kadian, Al-Dahle, Letman, Mathur, Schelten, Yang, Fan et~al.}]{llama3}
Abhimanyu Dubey, Abhinav Jauhri, Abhinav Pandey, Abhishek Kadian, Ahmad Al-Dahle, Aiesha Letman, Akhil Mathur, Alan Schelten, Amy Yang, Angela Fan, et~al. 2024.
\newblock The llama 3 herd of models.
\newblock \emph{arXiv preprint arXiv:2407.21783}.

\bibitem[{Galassi et~al.(2020)Galassi, Lippi, and Torroni}]{attention}
Andrea Galassi, Marco Lippi, and Paolo Torroni. 2020.
\newblock Attention in natural language processing.
\newblock \emph{IEEE transactions on neural networks and learning systems}, 32(10):4291--4308.

\bibitem[{Gan et~al.(2021{\natexlab{a}})Gan, Chen, Huang, Purver, Woodward, Xie, and Huang}]{syn}
Yujian Gan, Xinyun Chen, Qiuping Huang, Matthew Purver, John~R. Woodward, Jinxia Xie, and Pengsheng Huang. 2021{\natexlab{a}}.
\newblock \href {https://doi.org/10.18653/v1/2021.acl-long.195} {Towards robustness of text-to-{SQL} models against synonym substitution}.
\newblock In \emph{Proceedings of the 59th Annual Meeting of the Association for Computational Linguistics and the 11th International Joint Conference on Natural Language Processing (Volume 1: Long Papers)}, pages 2505--2515, Online. Association for Computational Linguistics.

\bibitem[{Gan et~al.(2021{\natexlab{b}})Gan, Chen, and Purver}]{dk}
Yujian Gan, Xinyun Chen, and Matthew Purver. 2021{\natexlab{b}}.
\newblock \href {https://doi.org/10.18653/v1/2021.emnlp-main.702} {Exploring underexplored limitations of cross-domain text-to-{SQL} generalization}.
\newblock In \emph{Proceedings of the 2021 Conference on Empirical Methods in Natural Language Processing}, pages 8926--8931, Online and Punta Cana, Dominican Republic. Association for Computational Linguistics.

\bibitem[{Gao et~al.(2024)Gao, Wang, Li, Sun, Qian, Ding, and Zhou}]{DAIL}
Dawei Gao, Haibin Wang, Yaliang Li, Xiuyu Sun, Yichen Qian, Bolin Ding, and Jingren Zhou. 2024.
\newblock \href {https://doi.org/10.14778/3641204.3641221} {Text-to-sql empowered by large language models: A benchmark evaluation}.
\newblock \emph{Proc. VLDB Endow.}, 17(5):1132–1145.

\bibitem[{Guo et~al.(2024)Guo, Tian, Tang, Wang, Wen, Yang, and Wang}]{GPT-3.5}
Chunxi Guo, Zhiliang Tian, Jintao Tang, Pancheng Wang, Zhihua Wen, Kang Yang, and Ting Wang. 2024.
\newblock Prompting gpt-3.5 for text-to-sql with de-semanticization and skeleton retrieval.
\newblock In \emph{PRICAI 2023: Trends in Artificial Intelligence}, pages 262--274, Singapore. Springer Nature Singapore.

\bibitem[{Guo et~al.(2019)Guo, Zhan, Gao, Xiao, Lou, Liu, and Zhang}]{old_schema_linking1}
Jiaqi Guo, Zecheng Zhan, Yan Gao, Yan Xiao, Jian-Guang Lou, Ting Liu, and Dongmei Zhang. 2019.
\newblock \href {https://doi.org/10.18653/v1/P19-1444} {Towards complex text-to-{SQL} in cross-domain database with intermediate representation}.
\newblock In \emph{Proceedings of the 57th Annual Meeting of the Association for Computational Linguistics}, pages 4524--4535, Florence, Italy. Association for Computational Linguistics.

\bibitem[{Honnibal and Montani(2017)}]{spacy}
Matthew Honnibal and Ines Montani. 2017.
\newblock {spaCy 2}: Natural language understanding with {B}loom embeddings, convolutional neural networks and incremental parsing.
\newblock To appear.

\bibitem[{Karpukhin et~al.(2020)Karpukhin, Oguz, Min, Lewis, Wu, Edunov, Chen, and Yih}]{dpr}
Vladimir Karpukhin, Barlas Oguz, Sewon Min, Patrick Lewis, Ledell Wu, Sergey Edunov, Danqi Chen, and Wen-tau Yih. 2020.
\newblock \href {https://doi.org/10.18653/v1/2020.emnlp-main.550} {Dense passage retrieval for open-domain question answering}.
\newblock In \emph{Proceedings of the 2020 Conference on Empirical Methods in Natural Language Processing (EMNLP)}, pages 6769--6781, Online. Association for Computational Linguistics.

\bibitem[{Lee et~al.(2025)Lee, Park, Kim, and Park}]{MCS}
Dongjun Lee, Choongwon Park, Jaehyuk Kim, and Heesoo Park. 2025.
\newblock \href {https://aclanthology.org/2025.coling-main.24/} {{MCS}-{SQL}: Leveraging multiple prompts and multiple-choice selection for text-to-{SQL} generation}.
\newblock In \emph{Proceedings of the 31st International Conference on Computational Linguistics}, pages 337--353, Abu Dhabi, UAE. Association for Computational Linguistics.

\bibitem[{Lei et~al.(2020)Lei, Wang, Ma, Gan, Lu, Kan, and Chua}]{schemalinking}
Wenqiang Lei, Weixin Wang, Zhixin Ma, Tian Gan, Wei Lu, Min-Yen Kan, and Tat-Seng Chua. 2020.
\newblock \href {https://doi.org/10.18653/v1/2020.emnlp-main.564} {Re-examining the role of schema linking in text-to-{SQL}}.
\newblock In \emph{Proceedings of the 2020 Conference on Empirical Methods in Natural Language Processing (EMNLP)}, pages 6943--6954, Online. Association for Computational Linguistics.

\bibitem[{Li et~al.(2023{\natexlab{a}})Li, Zhang, Li, and Chen}]{RESDSQL}
Haoyang Li, Jing Zhang, Cuiping Li, and Hong Chen. 2023{\natexlab{a}}.
\newblock Resdsql: Decoupling schema linking and skeleton parsing for text-to-sql.
\newblock In \emph{Proceedings of the AAAI Conference on Artificial Intelligence}, volume~37, pages 13067--13075.

\bibitem[{Li et~al.(2023{\natexlab{b}})Li, Hui, Cheng, Qin, Ma, Huo, Huang, Du, Si, and Li}]{graphix}
Jinyang Li, Binyuan Hui, Reynold Cheng, Bowen Qin, Chenhao Ma, Nan Huo, Fei Huang, Wenyu Du, Luo Si, and Yongbin Li. 2023{\natexlab{b}}.
\newblock Graphix-t5: Mixing pre-trained transformers with graph-aware layers for text-to-sql parsing.
\newblock In \emph{Proceedings of the AAAI Conference on Artificial Intelligence}, volume~37, pages 13076--13084.

\bibitem[{Li et~al.(2023{\natexlab{c}})Li, Lv, Yan, Lin, Zhu, Ni, Xie, Wang, and Qiu}]{udr}
Xiaonan Li, Kai Lv, Hang Yan, Tianyang Lin, Wei Zhu, Yuan Ni, Guotong Xie, Xiaoling Wang, and Xipeng Qiu. 2023{\natexlab{c}}.
\newblock \href {https://doi.org/10.18653/v1/2023.acl-long.256} {Unified demonstration retriever for in-context learning}.
\newblock In \emph{Proceedings of the 61st Annual Meeting of the Association for Computational Linguistics (Volume 1: Long Papers)}, pages 4644--4668, Toronto, Canada. Association for Computational Linguistics.

\bibitem[{Li et~al.(2024)Li, Wang, Zhao, Yang, Du, Hu, Zhang, Ye, Li, Zhao, and Mao}]{PET}
Zhishuai Li, Xiang Wang, Jingjing Zhao, Sun Yang, Guoqing Du, Xiaoru Hu, Bin Zhang, Yuxiao Ye, Ziyue Li, Rui Zhao, and Hangyu Mao. 2024.
\newblock \href {https://arxiv.org/abs/2403.09732} {Pet-sql: A prompt-enhanced two-round refinement of text-to-sql with cross-consistency}.
\newblock \emph{Preprint}, arXiv:2403.09732.

\bibitem[{Lin et~al.(2020)Lin, Socher, and Xiong}]{bridge}
Xi~Victoria Lin, Richard Socher, and Caiming Xiong. 2020.
\newblock \href {https://doi.org/10.18653/v1/2020.findings-emnlp.438} {Bridging textual and tabular data for cross-domain text-to-{SQL} semantic parsing}.
\newblock In \emph{Findings of the Association for Computational Linguistics: EMNLP 2020}, pages 4870--4888, Online. Association for Computational Linguistics.

\bibitem[{Liu et~al.(2019)Liu, Ott, Goyal, Du, Joshi, Chen, Levy, Lewis, Zettlemoyer, and Stoyanov}]{roberta}
Yinhan Liu, Myle Ott, Naman Goyal, Jingfei Du, Mandar Joshi, Danqi Chen, Omer Levy, Mike Lewis, Luke Zettlemoyer, and Veselin Stoyanov. 2019.
\newblock \href {https://arxiv.org/abs/1907.11692} {Roberta: A robustly optimized bert pretraining approach}.
\newblock \emph{Preprint}, arXiv:1907.11692.

\bibitem[{Luo et~al.(2024)Luo, Xu, Liu, Pasupat, and Kazemi}]{in-context}
Man Luo, Xin Xu, Yue Liu, Panupong Pasupat, and Mehran Kazemi. 2024.
\newblock In-context learning with retrieved demonstrations for language models: A survey.
\newblock \emph{arXiv preprint arXiv:2401.11624}.

\bibitem[{Nan et~al.(2023)Nan, Zhao, Zou, Ri, Tae, Zhang, Cohan, and Radev}]{Enhancing}
Linyong Nan, Yilun Zhao, Weijin Zou, Narutatsu Ri, Jaesung Tae, Ellen Zhang, Arman Cohan, and Dragomir Radev. 2023.
\newblock \href {https://doi.org/10.18653/v1/2023.findings-emnlp.996} {Enhancing text-to-{SQL} capabilities of large language models: A study on prompt design strategies}.
\newblock In \emph{Findings of the Association for Computational Linguistics: EMNLP 2023}, pages 14935--14956, Singapore. Association for Computational Linguistics.

\bibitem[{OpenAI et~al.(2024)OpenAI, Achiam, Adler, Agarwal, Ahmad, Akkaya, Aleman, Almeida, Altenschmidt, Altman, Anadkat, Avila, Babuschkin, Balaji, Balcom, Baltescu, Bao, Bavarian, Belgum et~al.}]{GPT-4}
OpenAI, Josh Achiam, Steven Adler, Sandhini Agarwal, Lama Ahmad, Ilge Akkaya, Florencia~Leoni Aleman, Diogo Almeida, Janko Altenschmidt, Sam Altman, Shyamal Anadkat, Red Avila, Igor Babuschkin, Suchir Balaji, Valerie Balcom, Paul Baltescu, Haiming Bao, Mohammad Bavarian, Jeff Belgum, et~al. 2024.
\newblock \href {https://arxiv.org/abs/2303.08774} {Gpt-4 technical report}.
\newblock \emph{Preprint}, arXiv:2303.08774.

\bibitem[{Pourreza and Rafiei(2023)}]{DIN}
Mohammadreza Pourreza and Davood Rafiei. 2023.
\newblock \href {https://proceedings.neurips.cc/paper_files/paper/2023/file/72223cc66f63ca1aa59edaec1b3670e6-Paper-Conference.pdf} {Din-sql: Decomposed in-context learning of text-to-sql with self-correction}.
\newblock In \emph{Advances in Neural Information Processing Systems}, volume~36, pages 36339--36348. Curran Associates, Inc.

\bibitem[{Qi et~al.(2022)Qi, Tang, He, Wan, Cheng, Zhou, Wang, Zhang, and Lin}]{RASAT}
Jiexing Qi, Jingyao Tang, Ziwei He, Xiangpeng Wan, Yu~Cheng, Chenghu Zhou, Xinbing Wang, Quanshi Zhang, and Zhouhan Lin. 2022.
\newblock \href {https://doi.org/10.18653/v1/2022.emnlp-main.211} {{RASAT}: Integrating relational structures into pretrained {S}eq2{S}eq model for text-to-{SQL}}.
\newblock In \emph{Proceedings of the 2022 Conference on Empirical Methods in Natural Language Processing}, pages 3215--3229, Abu Dhabi, United Arab Emirates. Association for Computational Linguistics.

\bibitem[{Rajkumar et~al.(2022)Rajkumar, Li, and Bahdanau}]{EvaluateText-to-SQL}
Nitarshan Rajkumar, Raymond Li, and Dzmitry Bahdanau. 2022.
\newblock \href {https://arxiv.org/abs/2204.00498} {Evaluating the text-to-sql capabilities of large language models}.
\newblock \emph{Preprint}, arXiv:2204.00498.

\bibitem[{Shaw et~al.(2018)Shaw, Uszkoreit, and Vaswani}]{relationtransformer}
Peter Shaw, Jakob Uszkoreit, and Ashish Vaswani. 2018.
\newblock \href {https://doi.org/10.18653/v1/N18-2074} {Self-attention with relative position representations}.
\newblock In \emph{Proceedings of the 2018 Conference of the North {A}merican Chapter of the Association for Computational Linguistics: Human Language Technologies, Volume 2 (Short Papers)}, pages 464--468, New Orleans, Louisiana. Association for Computational Linguistics.

\bibitem[{Shen et~al.(2024)Shen, Vougiouklis, Diao, Vyas, Ji, and Pan}]{ast}
Zhili Shen, Pavlos Vougiouklis, Chenxin Diao, Kaustubh Vyas, Yuanyi Ji, and Jeff~Z. Pan. 2024.
\newblock \href {https://doi.org/10.18653/v1/2024.emnlp-main.449} {Improving retrieval-augmented text-to-{SQL} with {AST}-based ranking and schema pruning}.
\newblock In \emph{Proceedings of the 2024 Conference on Empirical Methods in Natural Language Processing}, pages 7865--7879, Miami, Florida, USA. Association for Computational Linguistics.

\bibitem[{Tian et~al.(2023)Tian, Zhang, Ning, Li, Kummerfeld, and Zhang}]{editable}
Yuan Tian, Zheng Zhang, Zheng Ning, Toby Jia-Jun Li, Jonathan~K. Kummerfeld, and Tianyi Zhang. 2023.
\newblock \href {https://doi.org/10.18653/v1/2023.emnlp-main.1004} {Interactive text-to-{SQL} generation via editable step-by-step explanations}.
\newblock In \emph{Proceedings of the 2023 Conference on Empirical Methods in Natural Language Processing}, pages 16149--16166, Singapore. Association for Computational Linguistics.

\bibitem[{Wang et~al.(2020)Wang, Shin, Liu, Polozov, and Richardson}]{RAT-sql}
Bailin Wang, Richard Shin, Xiaodong Liu, Oleksandr Polozov, and Matthew Richardson. 2020.
\newblock \href {https://doi.org/10.18653/v1/2020.acl-main.677} {{RAT-SQL}: Relation-aware schema encoding and linking for text-to-{SQL} parsers}.
\newblock In \emph{Proceedings of the 58th Annual Meeting of the Association for Computational Linguistics}, pages 7567--7578, Online. Association for Computational Linguistics.

\bibitem[{Wei et~al.(2022)Wei, Tay, Bommasani, Raffel, Zoph, Borgeaud, Yogatama, Bosma, Zhou, Metzler, Chi, Hashimoto, Vinyals, Liang, Dean, and Fedus}]{emergent}
Jason Wei, Yi~Tay, Rishi Bommasani, Colin Raffel, Barret Zoph, Sebastian Borgeaud, Dani Yogatama, Maarten Bosma, Denny Zhou, Donald Metzler, Ed~H. Chi, Tatsunori Hashimoto, Oriol Vinyals, Percy Liang, Jeff Dean, and William Fedus. 2022.
\newblock \href {https://arxiv.org/abs/2206.07682} {Emergent abilities of large language models}.
\newblock \emph{Preprint}, arXiv:2206.07682.

\bibitem[{Yu et~al.(2018{\natexlab{a}})Yu, Yasunaga, Yang, Zhang, Wang, Li, and Radev}]{syntaxsqlnet}
Tao Yu, Michihiro Yasunaga, Kai Yang, Rui Zhang, Dongxu Wang, Zifan Li, and Dragomir Radev. 2018{\natexlab{a}}.
\newblock \href {https://doi.org/10.18653/v1/D18-1193} {{S}yntax{SQLN}et: Syntax tree networks for complex and cross-domain text-to-{SQL} task}.
\newblock In \emph{Proceedings of the 2018 Conference on Empirical Methods in Natural Language Processing}, pages 1653--1663, Brussels, Belgium. Association for Computational Linguistics.

\bibitem[{Yu et~al.(2018{\natexlab{b}})Yu, Zhang, Yang, Yasunaga, Wang, Li, Ma, Li, Yao, Roman, Zhang, and Radev}]{spider}
Tao Yu, Rui Zhang, Kai Yang, Michihiro Yasunaga, Dongxu Wang, Zifan Li, James Ma, Irene Li, Qingning Yao, Shanelle Roman, Zilin Zhang, and Dragomir Radev. 2018{\natexlab{b}}.
\newblock \href {https://doi.org/10.18653/v1/D18-1425} {{S}pider: A large-scale human-labeled dataset for complex and cross-domain semantic parsing and text-to-{SQL} task}.
\newblock In \emph{Proceedings of the 2018 Conference on Empirical Methods in Natural Language Processing}, pages 3911--3921, Brussels, Belgium. Association for Computational Linguistics.

\bibitem[{Zhang et~al.(2023{\natexlab{a}})Zhang, Cao, Chen, Xu, and Yu}]{ACT}
Hanchong Zhang, Ruisheng Cao, Lu~Chen, Hongshen Xu, and Kai Yu. 2023{\natexlab{a}}.
\newblock \href {https://doi.org/10.18653/v1/2023.findings-emnlp.227} {{ACT}-{SQL}: In-context learning for text-to-{SQL} with automatically-generated chain-of-thought}.
\newblock In \emph{Findings of the Association for Computational Linguistics: EMNLP 2023}, pages 3501--3532, Singapore. Association for Computational Linguistics.

\bibitem[{Zhang et~al.(2023{\natexlab{b}})Zhang, Lin, Wang, Zhang, Sun, Jianhe, Tan, Jiang, and Shen}]{REFSQL}
Kun Zhang, Xiexiong Lin, Yuanzhuo Wang, Xin Zhang, Fei Sun, Cen Jianhe, Hexiang Tan, Xuhui Jiang, and Huawei Shen. 2023{\natexlab{b}}.
\newblock \href {https://doi.org/10.18653/v1/2023.findings-emnlp.48} {{R}e{FSQL}: A retrieval-augmentation framework for text-to-{SQL} generation}.
\newblock In \emph{Findings of the Association for Computational Linguistics: EMNLP 2023}, pages 664--673, Singapore. Association for Computational Linguistics.

\bibitem[{Zhang et~al.(2022)Zhang, Zhang, Li, and Smola}]{automated-cot}
Zhuosheng Zhang, Aston Zhang, Mu~Li, and Alex Smola. 2022.
\newblock \href {https://arxiv.org/abs/2210.03493} {Automatic chain of thought prompting in large language models}.
\newblock \emph{Preprint}, arXiv:2210.03493.

\end{thebibliography}
